\documentclass[10pt,twocolumn,letterpaper]{article}
\usepackage[pagenumbers]{cvpr}
\usepackage{fontspec}
\usepackage{xeCJK}
\setCJKsansfont{FandolHei-Regular.otf}[BoldFont=FandolHei-Bold.otf,Script=Default]
\setCJKmonofont{FandolFang-Regular.otf}[Script=Default]
\usepackage{mathtools}
\usepackage{array,tabularx,multirow}
\usepackage{tikz}
\usetikzlibrary{arrows.meta,positioning,fit,backgrounds,calc}
\usepackage{enumitem}
\usepackage{placeins}
\usepackage{balance}
\definecolor{canvasblue}{HTML}{245B78}
\definecolor{stylegreen}{HTML}{3F735F}
\definecolor{glyphorange}{HTML}{A7572B}
\definecolor{draftgray}{HTML}{6E6E6E}
\newcommand{\method}{EditText}
\newcommand{\sg}{\operatorname{sg}}

\newcommand{\E}{\mathbb{E}}
\newcommand{\src}{\mathrm{src}}
\newcommand{\tgt}{\mathrm{tgt}}
\newcommand{\FM}{\mathrm{FM}}
\newcolumntype{Y}{>{\raggedright\arraybackslash}X}
\definecolor{cvprblue}{rgb}{0.21,0.49,0.74}
\usepackage[pagebackref,breaklinks,colorlinks,allcolors=cvprblue,unicode]{hyperref}

\def\confName{CVPR}
\def\confYear{2027}

\title{EditText: Structure-Aware Generation and Selective Visual Transport for Fine-Grained Text Editing}

\author{
  Zhaobin Chu$^{1,2,*}$ \qquad
  Guandong Li$^{2,*}$ \qquad
  Xin Lv$^{1,2}$ \qquad
  Zhiwei Zeng$^{2}$
  \\[0.5em]
  $^{1}$University of Science and Technology of China, Hefei, China\\
  $^{2}$iFlytek, Hefei, China\\
  {\ttfamily\small zbchu@mail.ustc.edu.cn}\\
  {\ttfamily\small gdli7@iflytek.com, xinlv@iflytek.com, zwzeng@iflytek.com}
  \\[0.3em]
  {\small $^{*}$These authors contributed equally to this work.}
}

\date{}

\hypersetup{
  pdftitle={EditText: Structure-Aware Generation and Selective Visual Transport for Fine-Grained Text Editing},
  pdfauthor={}
}

\makeatletter
\g@addto@macro\@maketitle{\begin{minipage}{\textwidth}
\centering
\includegraphics[width=.96\textwidth]{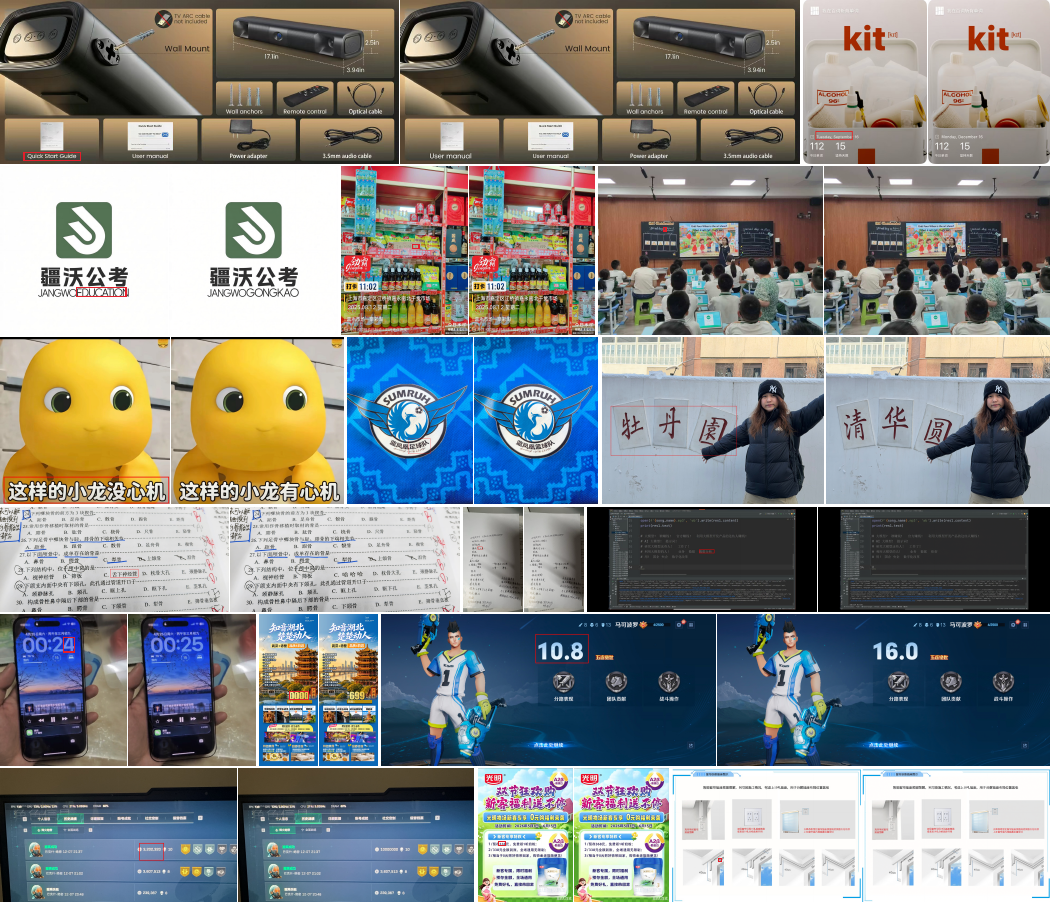}
\captionsetup{hypcap=false}
\captionof{figure}{%
\textbf{Fine-grained text editing across three text categories.}
From top to bottom, two rows each present Alphabetic, Logographic,
and Arabic numerals, for a total of 17 examples. Each pair shows
the source image with a red edit box on the left and the
corresponding result on the right.
Full images are tightly arranged at their original aspect ratios.
Alphabetic examples include phrase length changes, logo letters,
dense storefront text, and very small English text.
Logographic examples include outlined subtitles, curved team
emblems, scene plaques, dense body text, and very small handwritten
and screen text.
Arabic numeral examples include screen numerals, digit-count
changes, small text on posters, and very small dimension labels
in installation diagrams.
}
\label{fig:teaser}
\end{minipage}}
\makeatother

\begin{document}

\maketitle

\clearpage
\begin{abstract}
Fine-grained text editing aims to accurately replace text in a specified region while preserving the source text's font, color, and texture, as well as visual consistency in unedited regions.
We present \method{}, a unified framework for fine-grained text editing that combines target character generation with visual transport from the source image.
The model uses dual-condition Fill inputs: a masked image and a source text style reference, both sharing a spatial reference with the target image.
The masked image hides the original text and retains unedited content as a source canvas; the style reference supplies appearance cues that guide new character generation together with the target text.
We introduce an Attention Transport framework in which Source Weight describes the attention allocated to a condition and the Spatial Aggregation Kernel describes its internal spatial attention distribution.
Accurate transfer of source details depends on attention strength, the retention of details in condition features, and spatially accurate aggregation.
For selected non-edit queries, we therefore preserve the total attention weight assigned to the masked image while redistributing its attention toward spatially corresponding neighborhoods, enabling selective visual transport.
Multi-stage training, combined with a Glyph-aware Structural Loss on local target character structure, teaches the model to generate accurate text, preserve source text appearance, and retain unedited details.
We also introduce TBench, a fine-grained text editing benchmark covering Alphabetic, Logographic, and Arabic numerals, to evaluate character updates, appearance preservation, and the preservation of unedited details.
On TBench, \method{} achieves state-of-the-art non-edit-region SSIM (0.9769) and full-image LPIPS (0.0460).
\end{abstract}

\section{Introduction}
\label{sec:intro}
Text in images exhibits diverse character systems and visual forms.
Alphabetic text commonly appears in product labels, descriptive text, and scene signage.
Logographic text, exemplified by Chinese characters, has rich stroke and component structures.
Arabic numerals convey precise information such as times, prices, quantities, and dimensions.

These three categories pose distinct editing challenges.
For Alphabetic text, changes in word or phrase length require adjustments to character spacing and spatial extent while retaining the original font and layout style.
For Logographic text, complex relationships among strokes and components place greater demands on local character structure.
For Arabic numerals, changes in values and digit counts require accurate numerical content while preserving font size, alignment, and visual style.
Dense layouts and very small text compound these challenges: the model must update fine character structures within a limited region while preserving adjacent text and background details.

Fine-grained text editing thus requires accurate character updates, source text appearance preservation, and the retention of unedited details across text types, lengths, and scales.
Figure~\ref{fig:teaser} shows our results across these three categories, including length changes, dense layouts, and very small text.
The method updates target content while retaining source text appearance and surrounding details.

\begin{figure}[t]
\centering
\includegraphics[width=\linewidth]{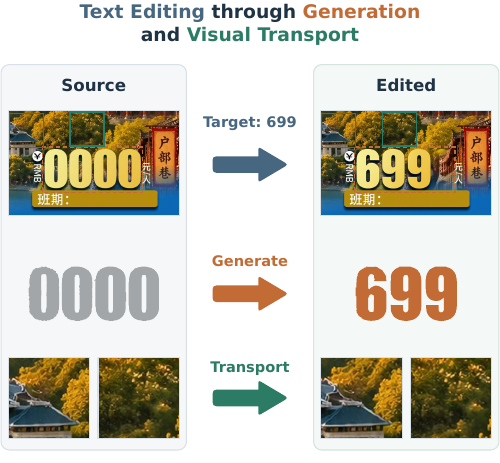}
\caption{\textbf{Text Editing through Generation and Visual Transport.}
Inside the orange dashed box, the model replaces ``0000'' with ``699'' in the original lettering style.
The green boxes highlight roof and foliage details that the model preserves by reusing visual features from the masked source image.
}
\label{fig:generate-transport-concept}
\end{figure}

\begin{figure*}[t]
\centering
\includegraphics[width=\textwidth]{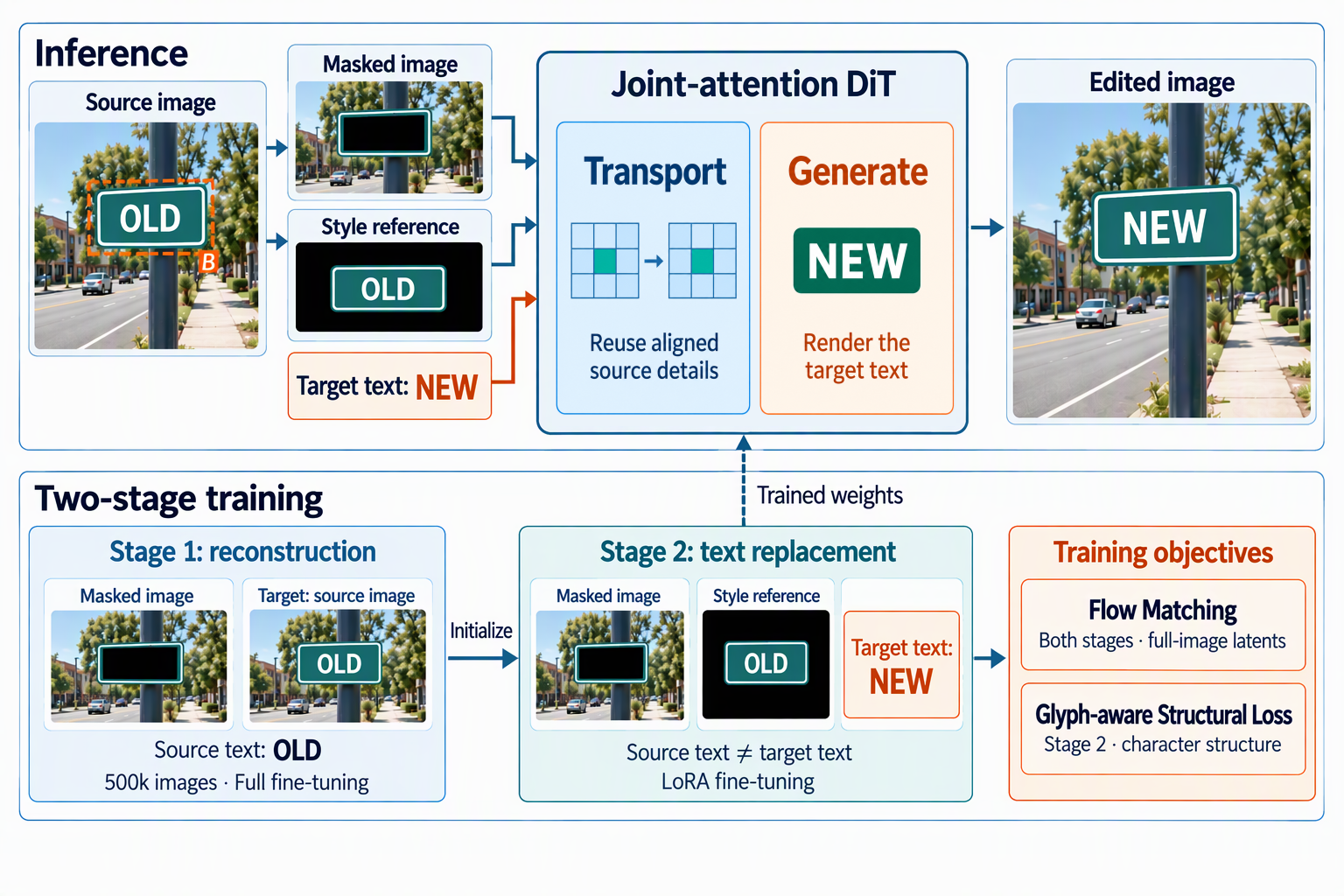}
\caption{\textbf{Overview of \method{}.}
Top: dual-condition Fill retains unedited content in the masked image and provides appearance cues through a source text style reference.
Both visual conditions share a spatial reference with the target image and guide a joint-attention DiT together with the target text.
For selected non-edit queries, spatial redistribution preserves the total attention weight $\mu_i$ assigned to the masked image while changing its internal spatial distribution from $\pi_{ij}$ to $\pi'_{ij}$.
Bottom: stage 1 performs self-supervised masked-text reconstruction and full-parameter fine-tuning on 500,000 real images. Stage 2 adds the style reference and learns text replacement through LoRA.
Both stages use Flow Matching supervision; stage 2 additionally constrains target character structure with the Glyph-aware Structural Loss.}
\label{fig:overview}
\end{figure*}

Diffusion models and multimodal generative Transformers have advanced image synthesis and editing~\cite{rombach2022latent,peebles2023dit,esser2024scaling}.
Text-aware methods further improve text generation through layout, glyph, and source text prompts~\cite{chen2023textdiffuser,tuo2023anytext,tuo2024anytext2,li2026selfprompt}.
Fine-grained editing across these three text categories requires selective use of source information: unedited content should remain intact, source text appearance should persist, and original character structures should be updated according to the target text.

To address these challenges, we propose \method{}, a unified framework for text editing through generation and visual transport (Fig.~\ref{fig:generate-transport-concept}).
Dual-condition Fill organizes character generation and source information reuse: the masked image hides the original text while retaining unedited content, and the source text style reference supplies appearance information from the masked region.
Both visual conditions share a spatial reference with the target image and jointly guide generation with the target text.
Within the edit region, the target text drives new character generation and the style reference guides appearance. In unedited regions, the model reads and reuses source information from the masked image.
Both functions operate within the same generative backbone.

Using a condition solely as a semantic or appearance reference does not necessarily preserve local information that can be reused by position.
Strong downsampling can weaken fine structures and textures, while missing spatial correspondence limits feature retrieval at matching locations.
We treat the masked image as a source canvas, emphasizing source detail retention and spatial alignment with the target image to support local information reuse in unedited regions.
To analyze how the source canvas is read, we introduce an Attention Transport framework:
Source Weight measures the attention allocated to a condition, and the Spatial Aggregation Kernel describes the spatial attention distribution within it.
Attention strength alone does not determine whether source details are transferred accurately; the condition features must retain those details, and aggregation must occur at the correct spatial locations.
Based on this observation, we preserve the total attention weight assigned to the masked image for selected non-edit queries while redistributing its internal attention toward corresponding spatial neighborhoods, yielding Selective Visual Transport.

Character structures in the edit region must be regenerated according to the target text while retaining source text appearance.
To learn character generation and unedited content preservation, we first perform self-supervised Fill reconstruction and full-parameter fine-tuning on 500,000 real images, using each original image as the target to restore masked text and preserve its surroundings.
We then add the source text style reference and perform text-replacement supervised fine-tuning (SFT) on examples with different source and target strings.
This stage learns new character generation and appearance preservation, with a Glyph-aware Structural Loss strengthening local character structure supervision alongside the full-image objective.

Our main contributions are as follows:
\begin{itemize}
\item \textbf{We introduce \method{} for fine-grained text editing.}
The framework combines target character generation with visual transport from the source image, using a source canvas and source text style reference to coordinate content updates and appearance preservation.
It combines self-supervised reconstruction and text replacement in two training stages, with a Glyph-aware Structural Loss that strengthens local character structure learning.
\item \textbf{We develop the Attention Transport framework and propose selective visual transport.}
Decomposing attention to the source condition into Source Weight and a Spatial Aggregation Kernel reveals that source detail transfer depends on both detail retention in condition features and spatially accurate retrieval.
For selected non-edit queries, spatial redistribution strengthens retrieval from corresponding neighborhoods while preserving the total attention weight assigned to the source canvas, enabling Selective Visual Transport.
\item \textbf{We achieve fine-grained editing that combines text updates, appearance preservation, and full-image detail preservation.}
The method supports Alphabetic, Logographic, and Arabic numerals, including length changes, dense layouts, very small text, and complex Logographic character structures.
It preserves unedited details during full-image generation, achieving state-of-the-art non-edit-region SSIM (0.9769) and full-image LPIPS (0.0460) on TBench.
\item \textbf{We introduce TBench, the first benchmark for fine-grained text editing.}
It covers Alphabetic, Logographic, and Arabic numerals, along with length changes, dense layouts, and very small text, and systematically evaluates character updates, text appearance preservation, and the preservation of unedited details.
\end{itemize}

\section{Related Work}
\label{sec:related}

\paragraph{Text generation and editing.}
TextDiffuser combines text layout with diffusion-based generation~\cite{chen2023textdiffuser}.
AnyText uses glyphs, positions, and masked images for multilingual text generation and editing, while AnyText2 adds font and color control~\cite{tuo2023anytext,tuo2024anytext2}.
Foundation models such as Qwen-Image and Seedream 3.0 support Chinese text generation and improve complex glyph rendering and text layout~\cite{wu2025qwen,gao2025seedream}.
Qwen-Image-Edit extends these capabilities to Chinese text editing~\cite{qwen2025edit}.
SelfPromptSTE builds style prompts from source text, provides target structure through rendered glyphs, and enables open-vocabulary editing through in-context learning~\cite{li2026selfprompt}.
\method{} uses dual-condition Fill: a masked image and a source text style reference share a spatial reference and provide a source canvas and text appearance cues, respectively. Target text and text-replacement supervision guide new character generation.

\paragraph{Image inpainting and instruction-based editing.}
Diffusion probabilistic models and score-based modeling provide foundations for image generation~\cite{ho2020ddpm,song2021score}, while latent diffusion enables high-resolution synthesis and editing~\cite{rombach2022latent}.
Palette studies diffusion-based image-to-image translation~\cite{saharia2022palette}.
SDEdit, DiffEdit, and RePaint enable editing or inpainting through stochastic differential equations, mask guidance, and known image content, respectively~\cite{meng2022sdedit,couairon2023diffedit,lugmayr2022repaint}.
From pix2pix to InstructPix2Pix, paired supervision has also been widely used to learn image transformations~\cite{isola2017pix2pix,brooks2023instructpix2pix}.
Fine-grained text editing additionally requires accurate generation of discrete characters and preservation of text appearance.
TextStyleBrush transfers visual text style from a single example~\cite{krishnan2023textstylebrush}.
\method{} preserves unedited content and introduces a source text style reference to provide direct appearance information for the masked region.

\paragraph{Attention interactions and visual preservation.}
DiT processes image latent tokens with Transformers~\cite{peebles2023dit}, and multimodal flow models use joint attention for bidirectional interactions between image and text tokens~\cite{esser2024scaling}.
ControlNet and T2I-Adapter improve spatial control through conditioning branches or adapters~\cite{zhang2023controlnet,mou2023t2iadapter}.
Plug-and-Play Diffusion Features uses intermediate features to preserve consistency during image editing~\cite{tumanyan2023plugplay}, and Prompt-to-Prompt demonstrates the effects of attention control on editing~\cite{hertz2023prompt}.
\method{} analyzes condition feature aggregation in terms of attention strength and spatial attention distribution, examines how spatial correspondence and condition representation quality affect visual preservation, and redistributes spatial attention within the masked image.

\section{Method}
\label{sec:method}
\subsection{Task Definition}
\label{sec:fill}
Given a source image $I_\src\in\mathbb R^{H\times W\times3}$, an edit region $B$, and a target string $y$, we formulate fine-grained text editing as
\begin{equation}
 \hat I=\mathcal F_\theta(I_\src,B,y).
 \label{eq:task}
\end{equation}
The edit region $B$ is specified by an annotation or a user-drawn box.
The model must generate the target string $y$, retain source text appearance, and preserve visual consistency in unedited regions.
Within a single generative process, \method{} generates target characters while retaining source text appearance in the edit region, and transports source visual information to preserve content in unedited regions.

Let $\Omega$ denote the full image domain. The edit and non-edit regions are
\begin{equation}
 \Omega_e=B,\qquad \Omega_n=\Omega\setminus B.
 \label{eq:regions}
\end{equation}
$\Omega_e$ is defined by an edit box or mask and includes both the text to be replaced and background pixels within that region.
$\Omega_n$ contains the rest of the image, including other text, objects, and background.
The model generates target characters and retains source text appearance in $\Omega_e$, while preserving source content in $\Omega_n$. Joint generation maintains visual continuity between the two regions.

We first describe the conditioning design and character structure supervision for character generation, then the source information retrieval mechanism for visual transport, and finally the staged training scheme that combines them.

\subsection{Character Generation and Appearance Preservation}
\label{sec:generate}
Character generation operates on the edit region $\Omega_e$, updating character structures according to the target text while using the font, color, and texture of the source text.
Dual-condition Fill provides complementary content and appearance information, and the Glyph-aware Structural Loss strengthens local character structure supervision.

\paragraph{Dual-condition Fill and the source text style reference.}
The target text specifies new character content, while the source text style reference provides appearance cues.
We organize the inputs in a Fill formulation. Let $M$ be the binary mask of $B$. Masking the text region produces the masked image.
To supply appearance information from the masked source text, we construct a style reference, yielding dual-condition Fill:
\begin{equation}
\begin{aligned}
 I_m&=(1-M)\odot I_\src+M\odot c,\\
 I_s&=M\odot I_\src+(1-M)\odot c,\\
 \hat I&=F_\theta(I_m,I_s,y),\qquad c=0.
\end{aligned}
\label{eq:fill}
\end{equation}
$I_m$ retains the content and positions of unedited regions, serving as a source canvas for visual information reuse.
$I_s$ retains the source text region at its original position on a black background, providing font, color, texture, and perspective references for new character generation.
Both visual conditions supplied to the VAE have the same dimensions and spatial layout as the target image and are aligned on the latent grid.
The target string is provided through a natural-language instruction, without an additional target glyph input.
The conditioning interface and coordinate processing are described in Appendices~\ref{app:glyph_input} and~\ref{app:region}, respectively.

The style reference $I_s$ also contains the strokes and components of the original characters, which cannot directly serve as target character structures.
We train on text-replacement examples with different source and target strings to learn appearance cues from $I_s$ and generate new strokes, components, and character arrangements according to $y$.
The two visual conditions and target text interact through joint attention. A single backbone generates the full image, connecting character updates with surrounding content during generation.

\paragraph{Full-image generation and local structure supervision.}
The model uses Flow Matching~\cite{lipman2023flow}.
Let $z_0=E(I_\tgt)$ be the target image latent. Noise interpolation and the target velocity are
\begin{equation}
 z_\sigma=(1-\sigma)z_0+\sigma\epsilon,\qquad
 v^*=\epsilon-z_0,
 \label{eq:flow}
\end{equation}
where $\epsilon\sim\mathcal N(0,\mathbf I)$. The base flow matching objective is
\begin{equation}
 \mathcal L_\FM=\E\!\left[w(\sigma)
    \operatorname{mean}[(v_\theta-v^*)^2]\right],
 \label{eq:fm}
\end{equation}
where $w(\sigma)$ follows the base scheduler.

After mapping the regions in Eq.~\eqref{eq:regions} to the latent grid, the per-sample mean squared term can be written as
\begin{equation}
 \operatorname{mean}[(v_\theta-v^*)^2]
 =\sum_{u\in\{e,n\}}\rho_u\ell_u,
 \quad\rho_u=|\Omega_u|/|\Omega|.
 \label{eq:regionfm}
\end{equation}
Here, $\ell_u$ is the regional mean squared error and $\rho_u$ is the area fraction. Training applies no regional reweighting.
The full-image mean squared objective aggregates errors in proportion to regional area. When the edit region is small, its character errors contribute to the overall objective in proportion to that area.
We therefore introduce a Glyph-aware Structural Loss during text-replacement training to provide additional local structure supervision for target characters.

\paragraph{Glyph-aware Structural Loss.}
Structural supervision acts on the predicted clean latent, rather than gradients of the velocity field or decoded RGB images.
From Eq.~\eqref{eq:flow},
\begin{equation}
 \hat z_0=z_\sigma-\sigma v_\theta.
 \label{eq:glyph-endpoint}
\end{equation}
We map the target text region to the latent grid and denote its cropping operator by $C_B$.
Let $D$ denote channel-wise horizontal and vertical Scharr gradients. We define
\begin{equation}
\begin{aligned}
 G_\theta&=D(C_B\hat z_0),\quad G_0=D(C_Bz_0),\\
 s_B&=\sg\!\left[\max\!\left(\sqrt{\operatorname{mean}(G_0^2)},s_{\min}\right)\right].
\end{aligned}
\label{eq:glyph-gradient}
\end{equation}
Here, $\sg$ denotes stop-gradient, and the mean covers spatial positions in the region, latent channels, and both gradient directions.
During training, a rasterized target glyph map undergoes the same transformations as the target image and is resized to nonnegative regional support weights $S(p)$.
This support map is used only in the loss and is not a model condition.
The per-sample structural error is
\begin{equation}
\begin{aligned}
 e_g(p)&=\sqrt{\left\|(G_\theta(p)-G_0(p))/s_B\right\|_2^2+\delta^2}-\delta,\\
 R_g&=\frac{\sum_p S(p)e_g(p)}{\max(\sum_p S(p),\varepsilon_S)}.
\end{aligned}
 \label{eq:glyph-raw}
\end{equation}
The norm covers channels and gradient directions, and $\delta$ is the Charbonnier smoothing constant.
The target image latent supplies structural supervision, while the glyph support map defines spatial weights. Supervision thus retains the target image's actual appearance rather than the pixel shape of a rendered font.

To limit the auxiliary term's influence on full-image reconstruction, let $\ell_\FM$ denote the per-sample flow matching loss. We apply noise gating and an adaptive amplitude coefficient:
\begin{equation}
\begin{aligned}
 g(\sigma)&=4\sigma(1-\sigma),\\
 c_g&=\sg\!\left[\min\!\left(1,
       \frac{\sg[\ell_\FM]}{\sg[R_g]+\varepsilon_R}\right)\right],\\
 \mathcal L_{\mathrm{glyph}}&=\E[g(\sigma)c_gR_g].
\end{aligned}
\label{eq:glyph-budget}
\end{equation}
The gate is larger at intermediate noise levels and approaches zero at both endpoints.
Let $\lambda_g\ge0$ be the structural loss coefficient; the full training objective is given in Sec.~\ref{sec:training}.
For nonnegative $\ell_\FM$, the auxiliary term satisfies $0\le\lambda_g g(\sigma)c_gR_g\le\lambda_g\sg[\ell_\FM]$, bounding its loss magnitude but not its gradient norm.
Implementation details are provided in Appendix~\ref{app:glyph-loss}.

\subsection{Visual Transport and Detail Preservation}
\label{sec:transport}
Visual transport operates on the non-edit region $\Omega_n$ to preserve surrounding text, textures, and background structures during full-image generation.
Although the Fill input retains this content, the network must still retrieve source information at the correct output positions; providing a source image condition alone does not ensure accurate detail reconstruction.

\begin{figure*}[t]
\centering
\includegraphics[width=\textwidth]{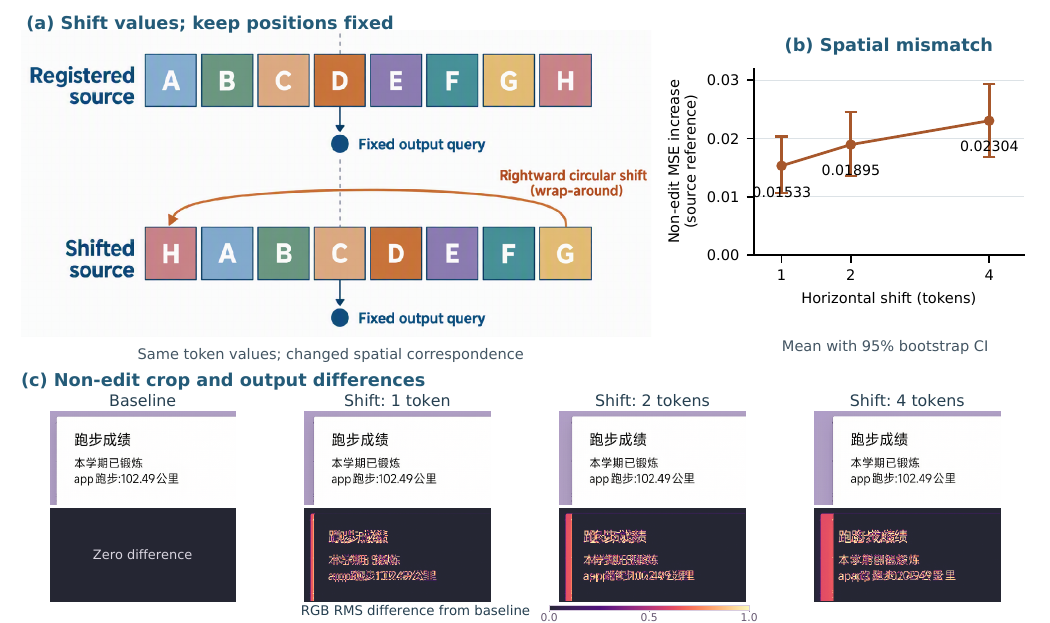}
\caption{\textbf{The role of spatial correspondence in preserving unedited details.}
Model weights, other inputs, and the random seed are fixed for a single-condition reference model.
(a) Source token values are circularly shifted horizontally while positional assignments and output queries remain fixed, associating the same query with different source content.
(b) Increase in source-relative non-edit-region MSE under condition latent shifts; error bars show image-wise bootstrap 95\% intervals. All three shifts degrade preservation.
(c) The same unedited region in all four outputs, with screenshot edges tinted lavender for display. The second row shows RGB root-mean-square differences from baseline, computed before tinting and plotted on a shared scale.
The curve measures source preservation error, while the difference maps show intervention-induced output changes.
}
\label{fig:transport-directional-misregistration}
\end{figure*}

\paragraph{Observing spatial retrieval in real images.}
Figure~\ref{fig:transport-directional-misregistration} compares outputs in unedited text regions.
Using a single-condition reference model, we fix the model weights, other inputs, and random seed, and circularly shift only the condition latents horizontally.
This preserves the set of source token values while changing their correspondence to output positions.
The aligned and shifted outputs retain similar overall layouts but exhibit local changes in text strokes and edges, whose locations are highlighted by difference maps with a shared color scale.
Across the evaluated images, shifts of 1, 2, and 4 tokens all increase source-relative error in unedited regions.
Source content availability alone is therefore insufficient for detail preservation; the model's use of spatial positions also matters.
Circular shifts additionally introduce boundary wraparound. Full settings and the scope of interpretation are given in Sec.~\ref{sec:causal} and Appendix~\ref{app:transport}.

\paragraph{From spatial correspondence to Attention Transport.}
These observations motivate two distinct questions: how much attention an output position allocates to the source image, and where that attention aggregates information within the source.
The former describes the source weight; the latter determines the spatial retrieval distribution.
Even with correct retrieval locations, details lost during condition encoding cannot be directly reused.
Attention Transport therefore analyzes source weights, spatial aggregation locations, and feature representation quality, motivating selective visual transport that preserves source weights.
Spatial misregistration tests the role of correspondence; component ablations of the full method evaluate the resulting editing gains (Table~\ref{tab:component-ablation}).

\paragraph{The source canvas and spatial correspondence.}
We treat the masked image as a source canvas, using a shared spatial reference to provide corresponding source neighborhoods for non-edit queries.
Attention Transport describes attention allocation and feature aggregation during this retrieval process.
Let $p_i,p_j$ be the image token coordinates of the output and condition, and let $R(p)$ denote rotary positional encoding.
For the relative-position form of RoPE~\cite{su2023roformer}, the attention score satisfies
\begin{equation}
 \langle R(p_i)q_i,R(p_j)k_j\rangle
 =q_i^\top R(p_j-p_i)k_j.
 \label{eq:rope}
\end{equation}
Relative displacement participates in query--key interactions, providing a positional reference for aggregating source features from corresponding neighborhoods.
Multi-image sequences also include non-spatial components such as image identity, so attention depends on both position and content.

\method{} applies consistent spatial transformations to both conditions and the target image to maintain coordinate correspondence.
Spatial alignment specifies retrieval locations, while detail reuse also depends on whether condition features retain source information.
The style reference retains source text positions and appearance; strokes of the original and target characters need not correspond one to one.

\begin{figure*}[t]
\centering
\includegraphics[width=\textwidth]{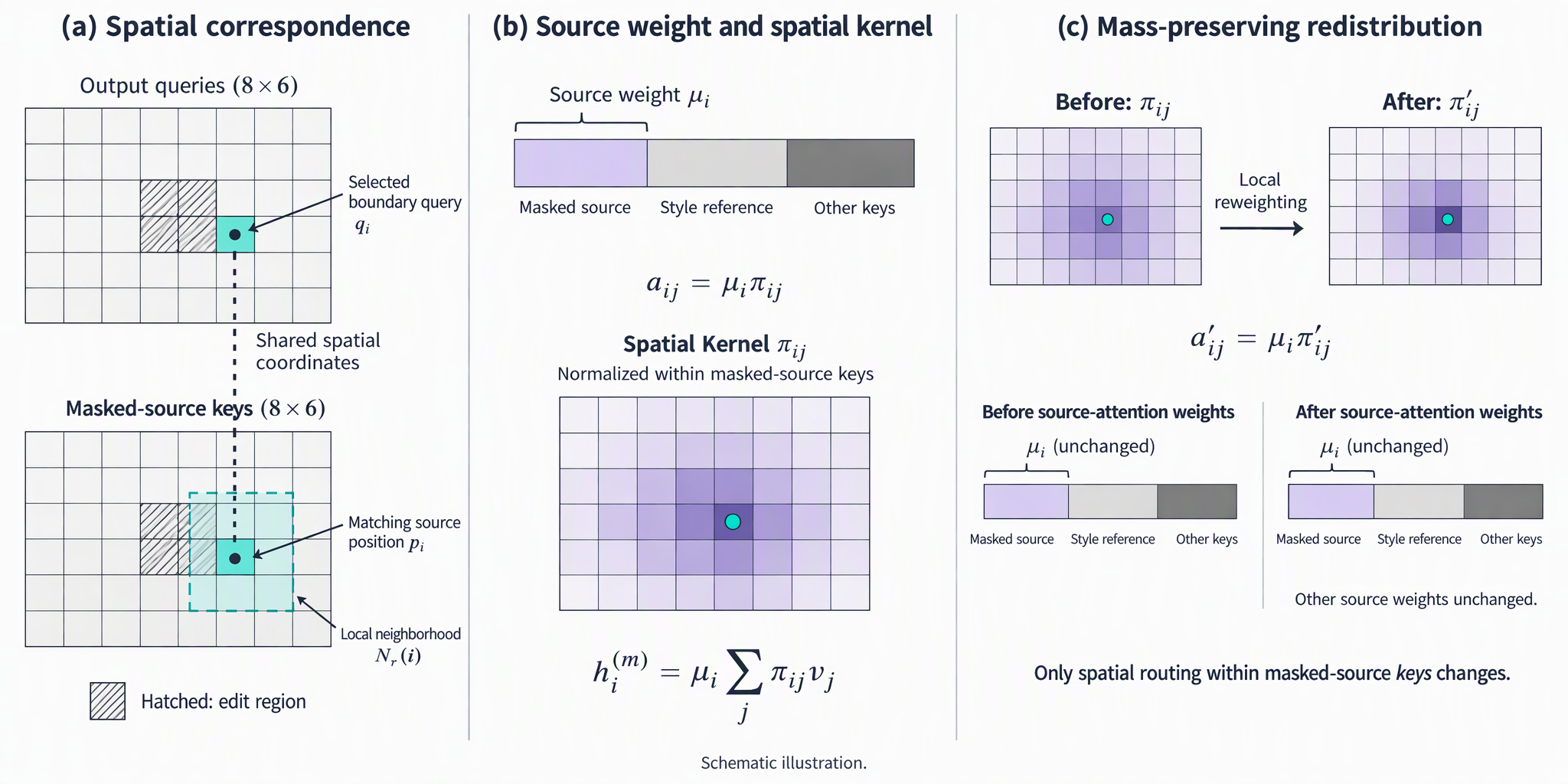}
\caption{\textbf{Attention Transport decomposition and spatial redistribution.}
(a) The masked image and target image share a spatial reference, associating each output query with a condition neighborhood $\mathcal N_r(i)$ at the same grid position.
(b) Attention to the masked image is decomposed into its total weight $\mu_i$ and an internally normalized spatial distribution $\pi_{ij}$.
(c) Redistribution adjusts the spatial attention distribution within the masked image while preserving $\mu_i$ and the attention weights of all other keys.
The grids schematically represent a single query's spatial attention distribution over the masked image.
}
\label{fig:attention-transport-overview}
\end{figure*}

\paragraph{Source Weight and Spatial Aggregation Kernel.}
Consider output query $i$ in one attention head at an arbitrary layer.
Let $a_{ij}$ be attention normalized over all visible keys and $\mathcal K_m$ the masked-image key set.
We define
\begin{equation}
\begin{aligned}
 \mu_i&=\sum_{j\in\mathcal K_m}a_{ij},\qquad
 \pi_{ij}=a_{ij}/\mu_i,\\
 h_i^{(m)}&=\mu_i\sum_{j\in\mathcal K_m}\pi_{ij}v_j.
\end{aligned}
\label{eq:transport}
\end{equation}
where $\mu_i>0$ and $\sum_j\pi_{ij}=1$.
The source weight $\mu_i$ sums the attention assigned by query $i$ to the masked image, characterizing the attention strength for that condition.
The spatial aggregation kernel $\pi_{ij}$ is the normalized spatial attention distribution within the condition.
The same decomposition applies to the style reference and requires no additional network branch.

For queries in $\Omega_n$, let the ideal source feature field $f(p)$ satisfy $\|f(p)-f(p')\|\le L\|p-p'\|$ within the analyzed range.
Let the condition values be $v_j=f(p_j)+e_j$, where $e_j$ denotes encoding and representation residuals. Then
\begin{equation}
\begin{aligned}
 \left\|\sum_j\pi_{ij}v_j-f(p_i)\right\|
 &\le L\,d_i+\bar e_i,\\
 d_i&=\sum_j\pi_{ij}\|p_j-p_i\|,\\
 \bar e_i&=\sum_j\pi_{ij}\|e_j\|.
\end{aligned}
\label{eq:localbound}
\end{equation}
This bound follows from the triangle inequality; its derivation is provided in Appendix~\ref{app:derivation}.
It contains a spatial mixing term $Ld_i$ and a condition representation residual $\bar e_i$, corresponding to aggregation distance and feature representation quality, respectively.
Under the stated assumptions, the bound characterizes normalized condition feature aggregation error.
The full output also depends on source weights, other conditions, subsequent network layers, and the decoder.

\paragraph{Selective Visual Transport.}
To strengthen retrieval from corresponding neighborhoods while retaining the source canvas's overall contribution, we apply mass-preserving spatial redistribution.
It fixes $\mu_i$ and concentrates attention within the masked image toward spatially corresponding neighborhoods.

For selected non-edit queries, let $\mathcal N_r(i)$ denote the local condition key neighborhood at the corresponding position, and set $b_{ij}=\beta\,\mathbf 1[j\in\mathcal N_r(i)]$, where $\beta\ge0$.
The redistribution is
\begin{equation}
\begin{aligned}
 \pi'_{ij}&=\frac{\pi_{ij}\exp(b_{ij})}
                  {\sum_{k\in\mathcal K_m}\pi_{ik}\exp(b_{ik})},\\
 a'_{ij}&=\mu_i\pi'_{ij},\qquad j\in\mathcal K_m .
\end{aligned}
\label{eq:redistribute}
\end{equation}
Thus, $\sum_{j\in\mathcal K_m}a'_{ij}=\mu_i$, the attention weights of all other keys remain unchanged, and $\beta=0$ recovers the original attention.
Local exponential weighting and renormalization make the spatial aggregation kernel independently adjustable.

\begin{figure*}[t]
\centering
\includegraphics[width=\textwidth]{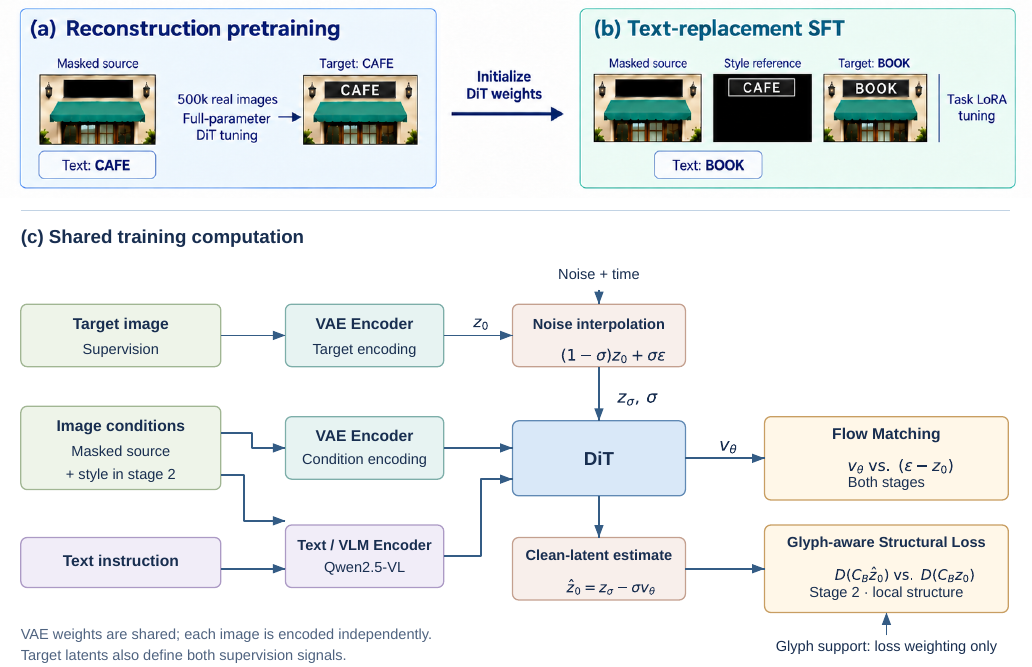}
\caption{\textbf{Two-stage training and shared computation.}
(a) Stage 1 fully fine-tunes the DiT through masked-text reconstruction on 500,000 real images.
(b) Stage 2 inherits these weights, adds the source text style reference, and learns text replacement through a task LoRA.
(c) The target image is VAE-encoded and interpolated with noise to form the DiT training input. Image conditions are encoded separately by a shared VAE and are also encoded jointly with the text instruction by Qwen2.5-VL.
Both stages supervise velocity prediction with Flow Matching. Stage 2 additionally recovers a predicted clean latent and compares its local structure in the edit region with that of the target latent.
$z_0$ is the target image latent; $C_B$ and $D$ denote edit-region cropping and Scharr gradients, respectively. Glyph support is used only to weight the structural loss.
}
\label{fig:two-stage-training}
\end{figure*}

\paragraph{Transport scope and division of roles.}
Spatial redistribution adjusts how selected non-edit queries read from the masked image, supporting content reuse from the source canvas.
Within the edit region, the model generates new strokes and arrangements according to $y$ and preserves source text appearance through the style reference.
Within the same backbone, these functions jointly replace text and preserve unedited content.

This yields two testable expectations: disrupting condition--output spatial correspondence should affect image preservation, and suppressing locally corresponding keys should alter the output.
Local attention aggregation can be characterized by the total attention weight $\mu_i$ assigned to the masked image, the mean aggregation distance $d_i$, and the local attention mass
\begin{equation}
 A_r=\frac{1}{|\mathcal Q_n|}\sum_{i\in\mathcal Q_n}
       \sum_{j\in\mathcal N_r(i)}a_{ij},
 \label{eq:mass}
\end{equation}
where $\mathcal Q_n$ is the set of non-edit queries.
$A_r$ depends on both the total masked-image attention weight and the fraction assigned to corresponding neighborhoods, and is interpreted jointly with $\mu_i$ and $d_i$.
Section~\ref{sec:causal} tests these expectations through condition aggregation features, input comparisons, and spatial interventions, and evaluates routing changes and image preservation gains separately.

\subsection{Multi-Stage Training}
\label{sec:training}
Both training stages use Fill to jointly learn character generation and unedited content preservation (Fig.~\ref{fig:two-stage-training}).
Stage 1 uses only the masked image as a visual condition and performs self-supervised reconstruction with the original text.
Stage 2 adds the source text style reference and performs text-replacement SFT with dual-condition Fill to learn character updates and appearance preservation.

\begin{figure*}[t]
\centering
\includegraphics[width=\textwidth]{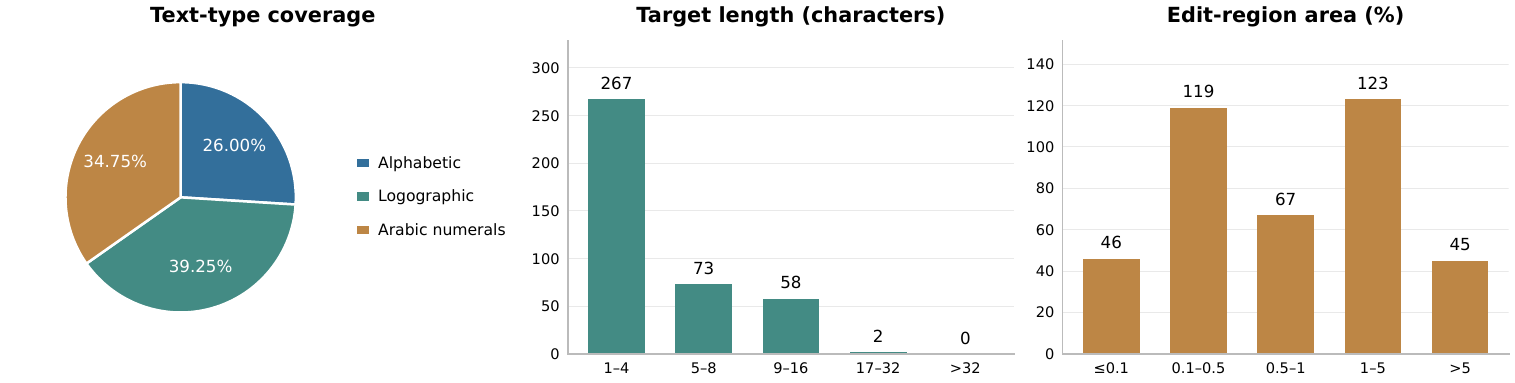}
\caption{\textbf{TBench data distributions.}
Left: proportions of the three text categories among all 400 tasks; each task belongs to exactly one category.
Middle: target string length in characters after whitespace removal. Right: edit-region area as a percentage of the source image area.
}
\label{fig:tbench-distribution}
\end{figure*}

\paragraph{Stage 1: Self-supervised text reconstruction.}
We construct reconstruction examples from 500,000 real images, using each original image as the target and its text-masked version as the sole visual condition.
The original text $y_\src$, obtained by OCR, is included in the instruction.
The task is
\begin{equation}
 \hat I^{(1)}=F_\theta(I_m,y_\src),\qquad I_\tgt=I_\src.
 \label{eq:stage1}
\end{equation}
This stage updates all DiT parameters to learn text-guided restoration of masked characters and preservation of source content in unedited regions.
Without a source text style reference, the model restores missing content from the text and surrounding image.
The resulting weights initialize subsequent text-replacement training.

\paragraph{Stage 2: Text-replacement SFT.}
Stage 2 inherits the reconstruction weights and adds the source text style reference in Eq.~\eqref{eq:fill}, extending reconstruction to text replacement.
Training examples with different source and target strings require the model to update character content while using the appearance reference.
The conditions, target image, and edit region undergo synchronized spatial transformations to preserve coordinate correspondence during training.

Stage 1 is initialized from Qwen-Image-Edit-2511~\cite{wu2025qwen,qwen2511card}, and stage 2 performs text-replacement SFT through LoRA~\cite{hu2022lora}.
Both stages use the full-image Flow Matching objective defined in Sec.~\ref{sec:generate} (Eq.~\eqref{eq:fm}).
Stage 1 sets $I_\tgt=I_\src$ to learn text restoration and unedited content preservation.
Stage 2 uses the edited image as the target and adds the Glyph-aware Structural Loss (Eq.~\eqref{eq:glyph-budget}) to full-image flow matching supervision:
\begin{equation}
 \mathcal L=\mathcal L_\FM+\lambda_g\mathcal L_{\mathrm{glyph}}.
 \label{eq:total}
\end{equation}
We set $\lambda_g=0.05$ in stage 2.

Both losses are averaged over samples in stage 2.

\section{TBench: A Benchmark for Fine-Grained Text Editing}
\label{sec:tbench-benchmark}

\subsection{Design Goals and Task Definition}
Evaluating fine-grained text editing requires assessing target text accuracy, source text appearance preservation, and unedited content preservation together.
Text recognition metrics do not fully capture consistency in font, color, and texture, while full-image similarity can obscure character errors in small regions.
We therefore construct TBench to evaluate text accuracy, visual preservation, and overall editing quality in a unified benchmark.
Each task provides a source image, an edit region, and a target string, and requires a complete edited image as output.
Evaluation assesses both text updates within the specified region and the preservation of surrounding text and background details.

\subsection{Dataset Construction and Coverage}
TBench includes screen interfaces, scene text, images combining graphics and text, receipts, and other scenarios.
Each task contains a source image, a target string, edit-region coordinates, and the annotated image dimensions.
During evaluation, edit regions are mapped according to image dimensions so that text recognition and non-edit-region evaluation use corresponding spatial regions.
Figure~\ref{fig:tbench-distribution} shows the distributions of text type, target string length, and edit-region area.

TBench covers Alphabetic, Logographic, and Arabic numerals, with 104, 157, and 139 tasks, respectively, accounting for 26.00\%, 39.25\%, and 34.75\% of all tasks.
The categories are mutually exclusive, with each task assigned to exactly one category.

Across these text types, TBench further covers length changes, dense layouts, and very small text.
Length changes test character arrangement within the original space and appearance preservation; dense layouts test targeted text updates and adjacent text preservation; very small text tests character structure and detail preservation within a limited pixel area.
The complex strokes and components of Logographic text place additional demands on local structural accuracy.

\subsection{Evaluation Protocol}
\label{sec:region_eval}
We evaluate text accuracy with PP-OCRv4. The annotated box is mapped to the output and expanded by 10 pixels. Local recognized text is concatenated and spaces are removed before computing target-string inclusion rate (OCR) and normalized Levenshtein distance (ED).
ED is normalized by the greater length of the recognized and target strings.
Image preservation is measured by non-edit-region SSIM~\cite{wang2004ssim} and full-image AlexNet LPIPS~\cite{zhang2018lpips}, which assess local structural preservation and overall perceptual difference, respectively.
Overall editing quality is assessed by Gemini-3.0-pro using a vision-language model score (LLM) from 1 to 5, with higher scores indicating better quality.
Human evaluation uses three labels: Good, Middle, and Bad.
OCR inclusion rate differs from exact full-string accuracy.
Reference images and image processing details are provided in Appendix~\ref{app:region}.

Text metrics assess content updates, and image preservation metrics assess source detail retention.
Overall visual scores and human evaluation further assess font, color, layout, and overall editing quality.
Together, these metrics characterize fine-grained editing beyond text recognition or full-image similarity alone.

\section{Experimental Results}
\label{sec:experiments}

\begin{table*}[tp]
\centering
\caption{Quantitative comparison on TBench.
The three column groups evaluate text accuracy, image preservation, and overall quality. SSIM is computed over non-edit regions and LPIPS over full images.
The 4-step and 8-step suffixes denote distilled variants with the corresponding sampling steps.
}
\label{tab:tbench-baseline}
\label{tab:tbench}
\small
\setlength{\tabcolsep}{4pt}
\renewcommand{\arraystretch}{1.08}
\begin{tabular*}{\textwidth}{@{\extracolsep{\fill}}lrrrrr@{}}
\toprule
& \multicolumn{2}{c}{Text accuracy} & \multicolumn{2}{c}{Image preservation} & \multicolumn{1}{c}{Overall quality}\\
\cmidrule(lr){2-3}\cmidrule(lr){4-5}\cmidrule(lr){6-6}
Model & OCR$\uparrow$ & ED$\downarrow$ & SSIM$\uparrow$ & LPIPS$\downarrow$ & LLM$\uparrow$\\
\midrule
FireRed Image Edit & .7200 & .2824 & .8968 & .0765 & 4.005\\
FireRed Image Edit (4-step) & .6970 & .3034 & .9432 & .0560 & 3.970\\
\addlinespace
Qwen-Image-Edit-2511 & .7121 & .3076 & .9392 & .0598 & 3.939\\
Qwen-Image-Edit-2511 (8-step) & .6633 & .3447 & .9494 & .0520 & 3.874\\
Qwen Image Edit Max & .6492 & .3598 & .9072 & .0869 & 3.806\\
Qwen Image Edit Plus & .6173 & .3692 & .8802 & .0796 & 3.653\\
Qwen Image 2 & .5990 & .4137 & .8916 & .0893 & 3.279\\
Qwen Image 2 Pro & \textbf{.7277} & .3330 & .8581 & .0777 & 3.162\\
Qwen AIO v1.8 & .3950 & .5499 & .6434 & .3951 & 3.145\\
\addlinespace
Seedream 4 & .4774 & .4275 & .7628 & .1757 & 2.598\\
Seedream 4.5 & .7020 & .2838 & .8552 & .1723 & 3.909\\
\addlinespace
Wan Image 2.6 & .6354 & .3561 & .8886 & .0763 & 3.667\\
\addlinespace
Nano Banana & .6450 & .3518 & .8596 & .0906 & \textbf{4.490}\\
Nano Banana 2 & .6889 & .3202 & .8595 & .0832 & 4.450\\
Nano Banana Flash & .3249 & .5842 & .7158 & .1847 & 2.893\\
\midrule
\textbf{\method{} (ours)} & .7093 & \textbf{.2352} & \textbf{.9769} & \textbf{.0460} & 4.1756\\
\bottomrule
\end{tabular*}
\end{table*}

\begin{table*}[tp]
\centering
\caption{Human evaluation pass rates on TBench.
The 4-step and 8-step suffixes denote distilled variants with the corresponding sampling steps.
}
\label{tab:human-review}
\small
\setlength{\tabcolsep}{3pt}
\renewcommand{\arraystretch}{1.08}
\begin{tabularx}{\textwidth}{@{}>{\raggedright\arraybackslash}X>{\raggedleft\arraybackslash}p{50pt}@{\hspace{20pt}}>{\raggedright\arraybackslash}X>{\raggedleft\arraybackslash}p{50pt}@{}}
\toprule
Model & Pass rate$\uparrow$ & Model & Pass rate$\uparrow$\\
\midrule
FireRed Image Edit & 45.50\% & Seedream 4 & 36.22\%\\
FireRed Image Edit (4-step) & 39.47\% & Seedream 4.5 & 57.07\%\\
Qwen-Image-Edit-2511 (8-step) & 44.22\% & Seedream 5 Lite & 44.33\%\\
Qwen Image Edit Max & 36.61\% & Seedream 5 Pro & 64.21\%\\
Qwen Image Edit Plus & 47.74\% & Wan Image 2.6 & 43.72\%\\
Qwen Image 2 & 34.36\% & Nano Banana Pro & 75.00\%\\
Qwen Image 2 Pro & 32.98\% & Nano Banana & 24.87\%\\
Qwen Image 3 Pro & 44.95\% & Nano Banana 2 & 70.95\%\\
Qwen AIO v1.8 & 21.88\% & SenseNova U1.5 & 47.00\%\\
GPT-image-2 & 57.07\% & gpt-image-2.5-flare & 54.04\%\\
gpt-image-2.5-sunburst & 68.79\% & \textbf{\method{} (ours)} & \textbf{76.68\%}\\
\bottomrule
\end{tabularx}
\end{table*}

\subsection{Experimental Setup}
\paragraph{Stage 1 full-parameter fine-tuning.}
Starting from Qwen-Image-Edit-2511, we fully fine-tune the DiT for 5 epochs on 500,000 self-supervised text reconstruction examples.
Training uses 4 nodes, a global batch size of 64, a peak learning rate of $10^{-5}$ with warmup and cosine decay, and a maximum training area of $1024^2$.
The resulting weights initialize stage 2. The full configuration is provided in Appendix~\ref{app:stage1}.

\paragraph{Stage 2 text-replacement SFT.}
Stage 2 inherits the reconstruction weights and trains a rank-128 task LoRA for the DiT on 21,061 text-replacement examples.
Each example provides a source image, an edit region, a target string, and an edited image, with different source and target strings.
Both visual conditions and the target image undergo consistent spatial transformations.
Target image latents provide flow matching supervision, and offline glyph support maps weight the structural loss.
Training uses a constant learning rate of $2\times10^{-4}$ for 5 epochs, with 2 dataset repetitions per epoch, a global batch size of 64, a maximum training area of 3,500,000 pixels, and a structural loss weight of $\lambda_g=0.05$.
Dataset construction, training settings, and interfaces are described in Appendix~\ref{app:stage2-interface}.

\paragraph{Inference.}
Inference combines the task LoRA and a four-step distillation LoRA, both with strength 1.
We use 4 sampling steps, guidance scale 1, and random seed 42.
The target areas of both the output and VAE conditions are approximately $2048^2$, preserving the source aspect ratio with dimensions rounded to the grid.

At inference, Eq.~\eqref{eq:redistribute} redistributes spatial attention within the masked image at denoising step 0, layer 25, across all heads for boundary queries in a four-token-wide band outside the edit box. Layers and steps are zero-indexed.
Depending on the longer side of the edit box on the condition grid, we use $(r,\beta)=(2,0.15),(4,0.25),(6,0.35)$; thresholds are given in Appendix~\ref{app:mechanism-settings}.
For each query and head, redistribution preserves the total masked-image attention weight and leaves all other key weights unchanged.
The implementation adds the difference between the redistributed and original weighted features to the native attention output before the shared output projection.
This operation introduces no trainable parameters or condition tokens, and $\beta=0$ recovers the original path.

\paragraph{Evaluation metrics.}
We use OCR, ED, SSIM, LPIPS, LLM scores, and human evaluation as defined in Sec.~\ref{sec:region_eval}.

\subsection{Quantitative Comparison}
\label{sec:tbench}
On TBench, \method{} achieves a local OCR inclusion rate of 70.93\%, ED of 0.2352, non-edit-region SSIM of 0.9769, full-image LPIPS of 0.0460, and an LLM score of 4.1756.
As shown in Table~\ref{tab:tbench-baseline}, \method{} obtains the lowest ED, highest non-edit-region SSIM, and lowest full-image LPIPS, indicating strong preservation of unedited visual content while updating text.
Its human evaluation pass rate reaches 76.68\%, outperforming the compared methods (Table~\ref{tab:human-review}).

\subsection{Qualitative Results}
\label{sec:qualitative}
\begin{figure*}[!tp]
\centering
\includegraphics[width=\textwidth]{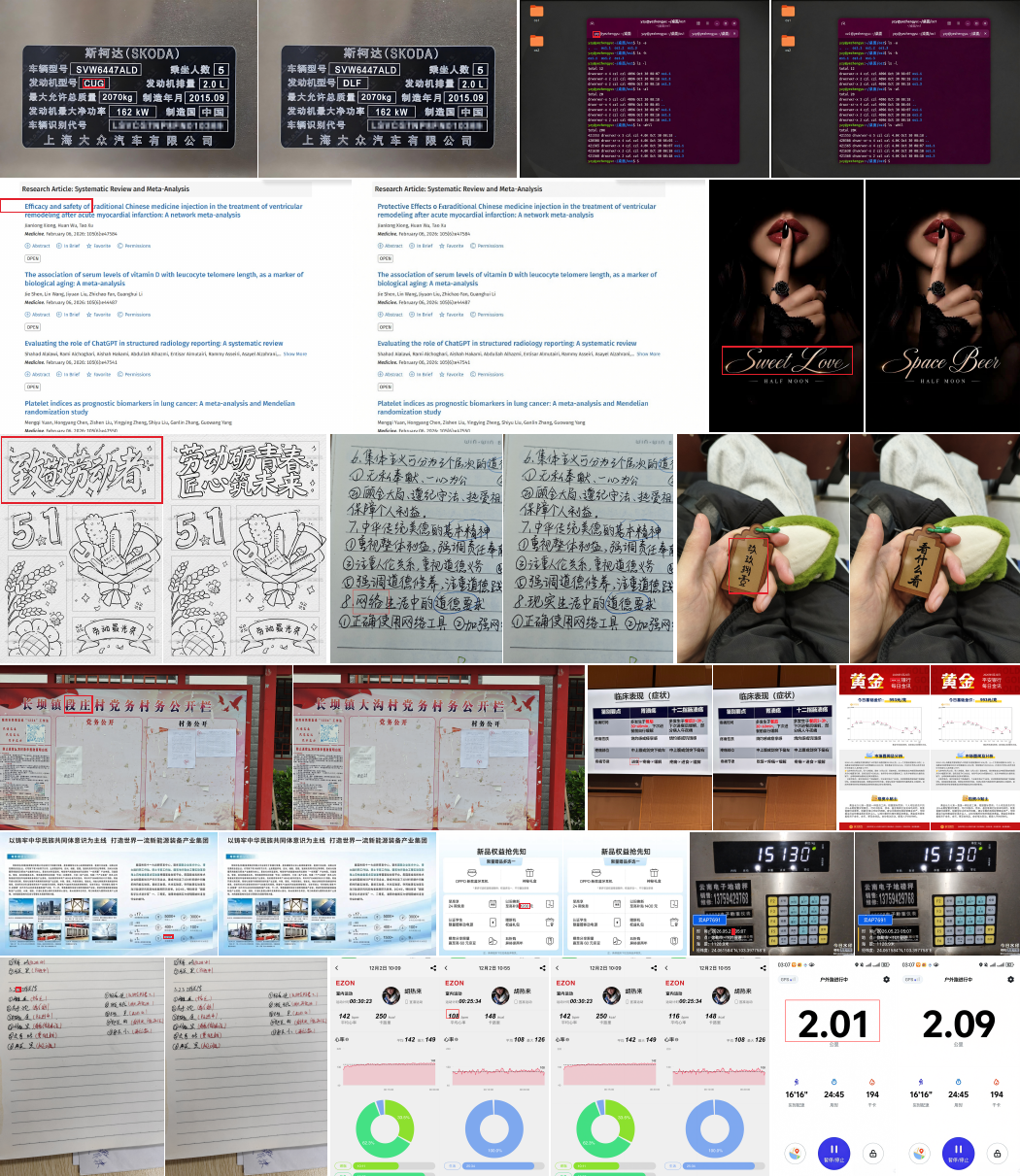}
\caption{\textbf{Fine-grained editing results across three text categories.}
TBench evaluation results, with two rows each for Alphabetic, Logographic, and Arabic numerals from top to bottom.
Each pair shows the source image on the left and the corresponding edited result on the right; red boxes indicate the edit regions in the source images.
Full images are tightly arranged while preserving their original aspect ratios.
}
\label{fig:qualitative-tbench}
\end{figure*}

Figure~\ref{fig:qualitative-tbench} presents fine-grained text editing results for Alphabetic, Logographic, and Arabic numerals, with two rows per category.
The examples show, respectively: letter updates on photographed nameplates and dense interfaces, length changes, and very small text; preservation of outlined and handwritten styles in complex Chinese characters and local updates in dense layouts; and changes in digit count, small numerals, and diverse strokes.
Full-image comparisons also show the preservation of adjacent text, textures, and background details in unedited regions.

\section{Ablation and Mechanism Analysis}
\label{sec:ablation-analysis}

\begin{figure*}[!tp]
\centering
\includegraphics[width=\textwidth]{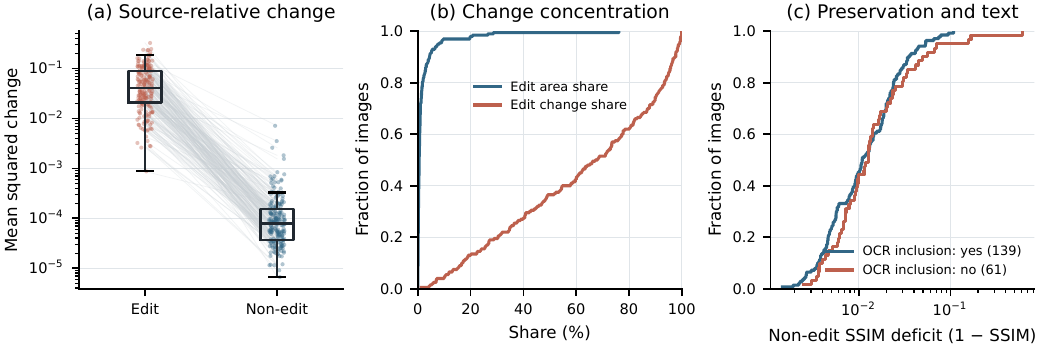}
\caption{\textbf{Change distribution and text correctness in dual-condition Fill.}
(a) Per-image comparison of source-relative mean squared changes in edit and non-edit regions; connected points correspond to the same image. Box plots show medians, quartiles, and 1.5 interquartile ranges.
(b) Edit-region area fraction and its share of full-image squared change.
(c) Cumulative distributions of non-edit-region $1-\mathrm{SSIM}$, grouped by OCR inclusion. The horizontal axis is logarithmic; smaller values indicate better preservation.
Squared change measures the magnitude of modifications from the source image.
}
\label{fig:region-change}
\end{figure*}

\begin{figure*}[!tp]
\centering
\includegraphics[width=.90\textwidth]{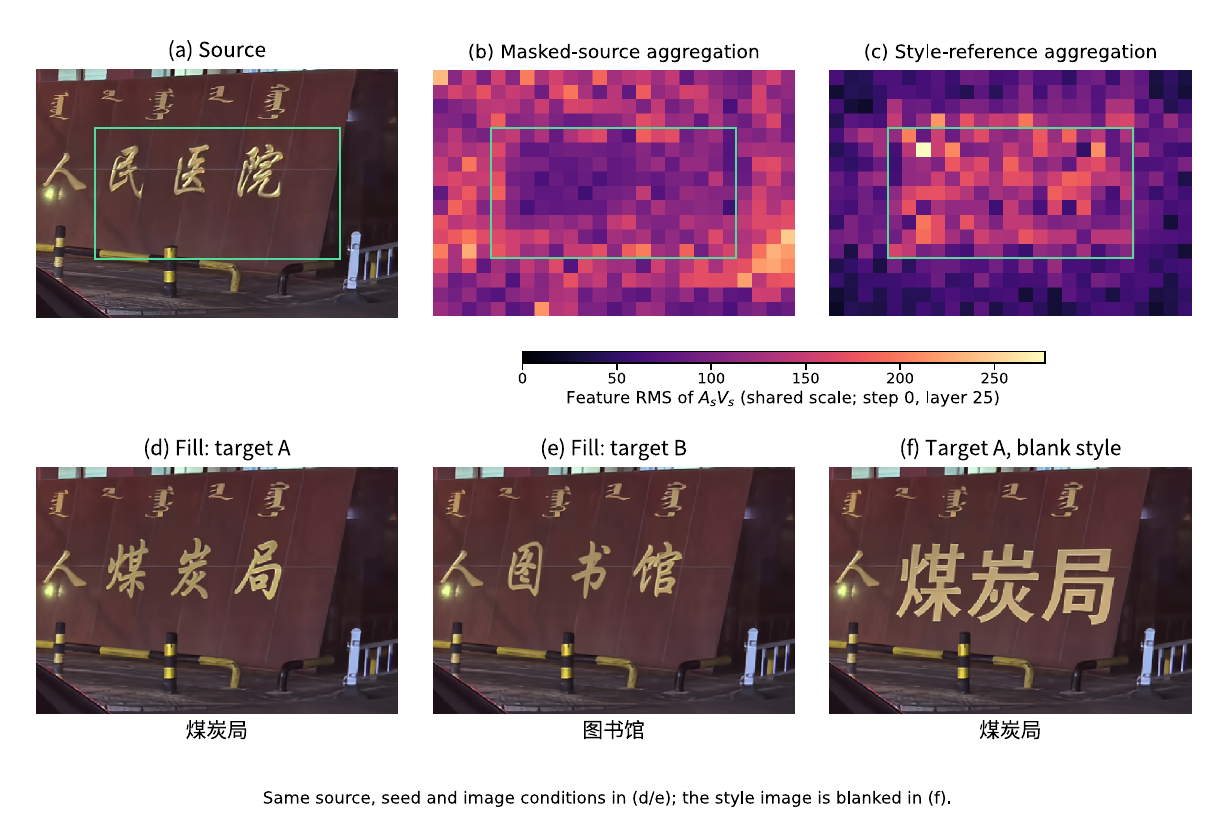}
\caption{\textbf{Condition aggregation and source text appearance reuse in Fill.}
(a) Source image crop; the green box marks the edit region on the token grid.
(b,c) RMS of the actual aggregated features $A_sV_s$ from the masked image and style reference at denoising step 0, layer 25. Both maps share a color scale; brighter values indicate larger feature magnitudes. Features are measured before the attention output projection and gating.
(d,e) Outputs with different target strings while fixing the source image, visual conditions, and random seed.
(f) The same target text as (d), with the style reference replaced by an equally sized black image.
Generated content changes with the target text. Blanking the style reference makes the text more regular and larger in this example.
The aggregation maps show weighted visual-condition feature responses; output comparisons assess their appearance effects.
}
\label{fig:fill-mechanism}
\end{figure*}

\begin{figure*}[!tp]
\centering
\includegraphics[width=.96\textwidth]{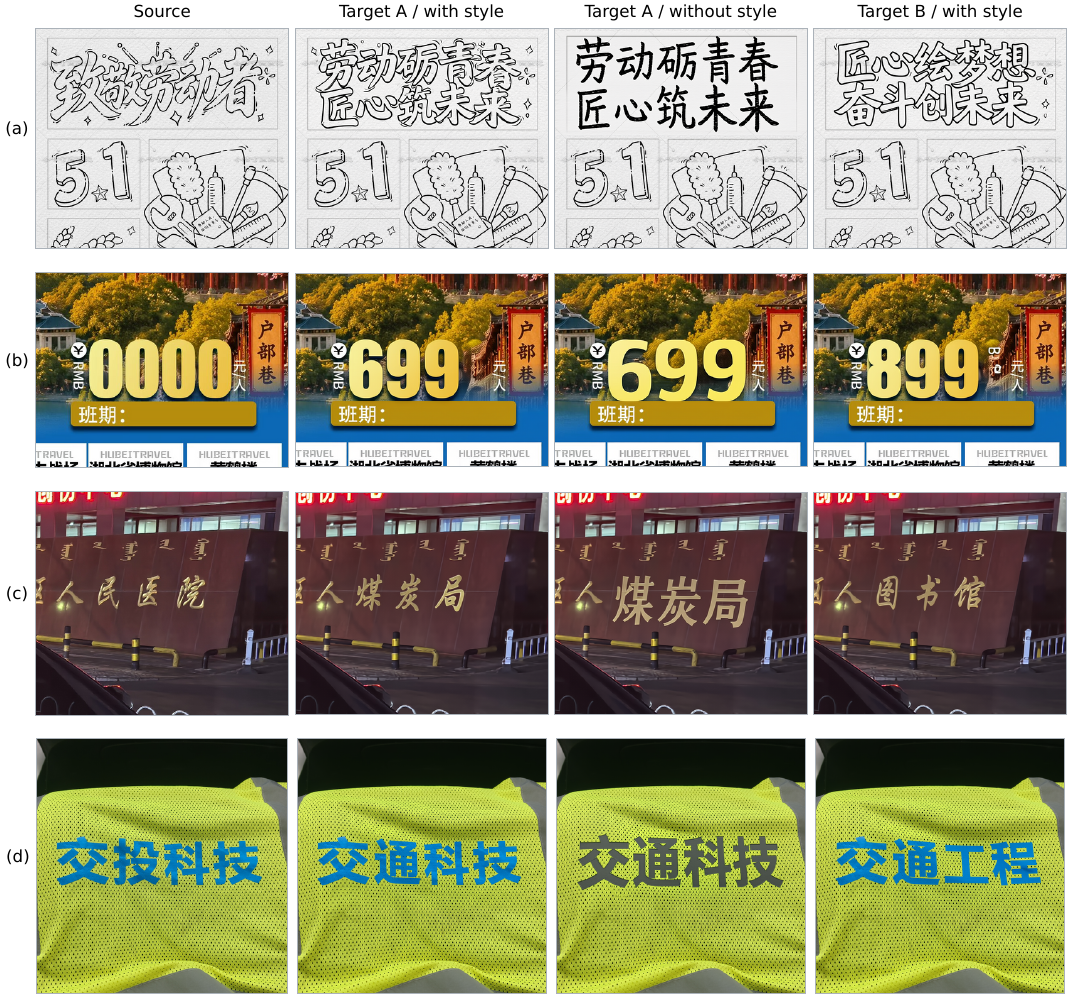}
\caption{\textbf{Target content updates and source text appearance preservation.}
Columns show the source image, target A with the style reference, target A with a blank style reference, and target B with the style reference.
Each row fixes the model, source image, edit box, and seed. Crops use the same spatial extent, retain their aspect ratios, and include surrounding image context.
Columns two and three change only the style reference; columns two and four change only the target text.
Targets A/B are: (a) 劳动砺青春 匠心筑未来 / 匠心绘梦想 奋斗创未来; (b) 699 / 899; (c) 煤炭局 / 图书馆; (d) 交通科技 / 交通工程.
}
\label{fig:fill-condition-roles}
\end{figure*}

\begin{figure*}[!tp]
\centering
\includegraphics[width=\textwidth]{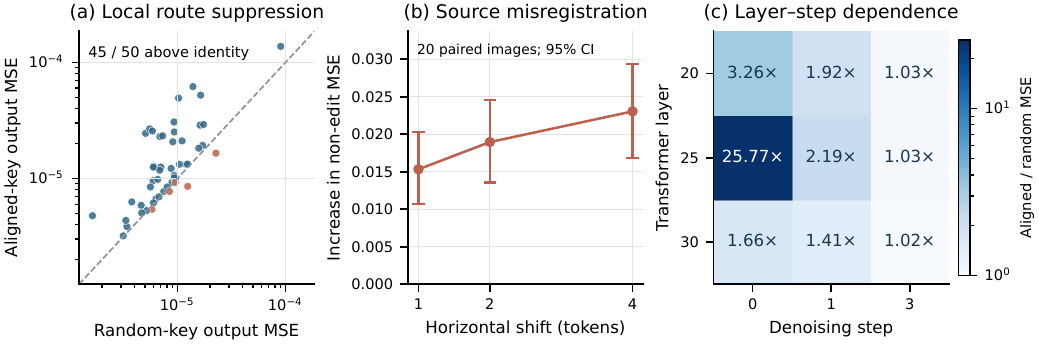}
\caption{\textbf{Attention Transport interventions with a single visual condition.}
(a) Corresponding-key and random-key suppression on the same image; MSE is measured relative to the unmodified output.
(b) Increase in source-relative non-edit-region MSE under condition latent shifts, with image-wise bootstrap 95\% intervals.
(c) Layer and denoising-step comparisons across nine experiments. Cell values give the ratio of mean perturbation MSE for corresponding-key suppression to that for random-key suppression.
}
\label{fig:transport_evidence}
\end{figure*}
We first analyze the contributions of the design components and training through component ablations and evaluation of reconstruction checkpoints.
We then examine information retrieval in Attention Transport through conditioning comparisons and spatial interventions.

\subsection{Component and Training Analysis}
\label{sec:ablation}
\paragraph{Component ablation.}
We remove one component at a time on TBench, using the same samples and evaluation settings for all ablations and keeping the remaining configuration fixed.
The full method includes self-supervised reconstruction initialization, the source text style reference, the Glyph-aware Structural Loss, and spatial attention redistribution.

\begin{table}[t]
\centering
\caption{Component ablation on TBench. The full method includes all four components.}
\label{tab:component-ablation}
\footnotesize
\setlength{\tabcolsep}{3pt}
\begin{tabular*}{\columnwidth}{@{\extracolsep{\fill}}lrrrrr@{}}
\toprule
Configuration & OCR$\uparrow$ & ED$\downarrow$ & SSIM$\uparrow$ & LPIPS$\downarrow$ & LLM$\uparrow$\\
\midrule
Full method & .7000 & .2059 & .9786 & .0452 & 4.125\\
w/o Stage 1 & .6600 & .2376 & .9789 & .0461 & 3.995\\
w/o style reference & .6700 & .2500 & .9627 & .0364 & 3.560\\
w/o glyph loss & .6950 & .2006 & .9753 & .0477 & 3.945\\
w/o redistribution & .6800 & .2441 & .9231 & .0501 & 3.765\\
\bottomrule
\end{tabular*}
\end{table}

Removing self-supervised reconstruction initialization reduces OCR from .7000 to .6600 and worsens ED and LLM scores, indicating that reconstruction training benefits subsequent text replacement.
Removing the style reference worsens OCR, ED, SSIM, and LLM scores but improves LPIPS. The source text reference thus affects multiple aspects of editing quality that a single perceptual distance does not fully capture.
Removing the structural loss worsens OCR, SSIM, LPIPS, and LLM scores while slightly improving ED, showing benefits across several metrics but not all text metrics.
Removing spatial redistribution reduces SSIM from .9786 to .9231 and worsens text metrics, LPIPS, and LLM scores, supporting the contribution of routing adjustment to editing quality.

\paragraph{Text replacement with reconstruction weights.}
\label{sec:stage1-eval}
We evaluate the reconstruction checkpoint at the end of the first epoch on text replacement (Table~\ref{tab:stage1-base}).
Training restores the original text, whereas evaluation supplies a new target string to assess transfer from reconstruction to text replacement.

\begin{table}[t]
\centering
\caption{Results of the first-epoch reconstruction checkpoint on TBench, using a single masked-image input and a general four-step acceleration LoRA.}
\label{tab:stage1-base}
\small
\setlength{\tabcolsep}{3pt}
\begin{tabular*}{\columnwidth}{@{\extracolsep{\fill}}lrrrrr@{}}
\toprule
Model & OCR$\uparrow$ & ED$\downarrow$ & SSIM$\uparrow$ & LPIPS$\downarrow$ & LLM$\uparrow$\\
\midrule
Stage 1 & .6650 & .2757 & .9634 & .0353 & 3.4600\\
\bottomrule
\end{tabular*}
\end{table}

The first-epoch reconstruction checkpoint achieves 66.50\% local OCR inclusion, 0.9634 non-edit-region SSIM, and 0.0353 full-image LPIPS.
Before text-replacement SFT, the model can already generate characters from new target text while retaining some source image content.
Stage 2 further introduces the source text appearance reference and learns target content replacement.

\paragraph{Reference model comparison.}
We compare two reference models with the same backbone on TBench.
The first trains a Fill LoRA on the original backbone, and the second trains a single-image editing LoRA on the reconstruction-fine-tuned backbone. Both are evaluated with a single image marked by a red box.
Their non-edit-region SSIM scores are 0.9750 and 0.9018, full-image LPIPS scores are 0.0468 and 0.1286, OCR inclusion rates are 0.6533 and 0.7588, and mean LLM scores are 3.9698 and 4.1658, respectively.
The first configuration preserves images better, while the second achieves higher text accuracy and overall scores.
The models also differ in training data and resolution, and this comparison illustrates quality trade-offs across configurations.
Paired intervals and configuration differences are provided in Appendix~\ref{app:comparison}.

\subsection{Attention Transport Mechanism Analysis}
\label{sec:causal}
Section~\ref{sec:transport} introduced source retrieval through spatial misregistration comparisons on real images.
Here, we provide the full experimental settings, quantitative statistics, and local routing interventions to analyze the roles of conditions and the effect of spatial correspondence on information transfer.
To study unedited content preservation, target text updates, and source text appearance preservation, we use a dual-condition Fill reference model trained by text-replacement SFT directly on the original backbone.
With the masked image and style reference as visual inputs, we change the target text, blank the style reference, and analyze aggregated features to examine how the model uses each input.
This reference model uses neither stage 1 reconstruction initialization nor the structural loss.

To further test spatial correspondence, we perform local key suppression and spatial misregistration experiments on an independently trained single-condition model.
Key suppression reduces pre-softmax attention scores at selected condition positions, weakening feature retrieval from those positions.
The single visual condition isolates local retrieval from the influence of a second condition image.
Each comparison changes only the factor under study while fixing model weights, other input conditions, and the random seed.
Detailed settings are provided in Appendices~\ref{app:implementation} and~\ref{app:transport}.

\paragraph{Change distribution and text correctness.}
Figure~\ref{fig:region-change}(a,b) summarizes pixel-wise squared changes between generated and source images.
Averaging per-image fractions, edit regions occupy only 2.12\% of image area but contribute 62.12\% of the full-image squared change, indicating that modifications concentrate in the replacement region.
These changes include removal of the original text and generation of new characters.
Grouping outputs by local OCR inclusion, the passing and failing groups achieve non-edit-region SSIM of 0.9829 and 0.9688, and full-image LPIPS of 0.0414 and 0.0538, respectively.
Outputs that fail OCR inclusion still preserve unedited regions well (Fig.~\ref{fig:region-change}(c)), supporting separate evaluation of text correctness and visual preservation.
Statistical definitions and coordinate processing are given in Appendix~\ref{app:change-analysis}.

\paragraph{Condition retrieval and appearance preservation in Fill.}
\label{sec:fill-pilot}
To analyze retrieval from the two visual conditions, we record the total attention weight and actual weighted feature aggregation for the masked image and style reference.
In Fig.~\ref{fig:fill-mechanism}(b,c), stronger masked-image responses appear around the edit region, while style-reference responses concentrate mainly within it.
At step 0, layer 25, the mean total attention weight assigned to the style reference is 10.37\%--13.31\% in the edit region and 5.91\%--6.79\% at the outer boundary, showing a consistent regional difference in attention allocation.
Per-sample values and attention collection settings are provided in Appendix~\ref{app:mechanism-settings}.
Spatial retrieval distributions within the masked image and corresponding-neighborhood statistics are provided in Appendix~\ref{app:feature-maps}.

\paragraph{Target content updates and source text appearance preservation.}
Figure~\ref{fig:fill-condition-roles} compares the effects of target text and style reference by changing the target string and blanking the reference.
With fixed visual conditions and a changed target string, output content updates while retaining the main appearance of the source text.
With fixed target text and a blank style reference, the hand-drawn heading changes from outlined to solid characters, poster numerals change from narrow and elongated to wide and rounded, sign text becomes more regular and larger, and printed fabric text changes from blue to dark gray-green.
These comparisons show how target text controls generated content and how the style reference influences new character appearance.
The reference effect varies across examples, and image context can also provide style cues.
Blanking the reference changes both VAE and VLM encodings, so this comparison assesses the style reference as a complete input.
Table~\ref{tab:component-ablation} evaluates the contribution of each component to overall quality.

\paragraph{The functional role of local condition routing.}
To test whether the model uses spatially corresponding source information when generating unedited regions, we compare two interventions in the single-condition setting.
One suppresses attention scores of condition keys in the neighborhood corresponding to each output position; the other suppresses the same number of random keys, using the same suppression strength.
Relative to the unmodified output, their mean perturbation MSE in unedited regions is $1.815\times10^{-5}$ and $1.028\times10^{-5}$, respectively.
A larger MSE indicates a greater output change when retrieval from those keys is weakened.
Corresponding-key suppression produces larger changes in 90\% of samples (Fig.~\ref{fig:transport_evidence}(a)).
Weakening retrieval from corresponding neighborhoods thus typically affects the output more than suppressing random positions, supporting the use of spatially corresponding source features in non-edit-region generation.
The two groups are not matched for total suppressed attention weight.
Settings and local output comparisons are provided in Appendix~\ref{app:transport}.

\paragraph{Spatial correspondence and layer--step dependence.}
Circularly shifting condition latents horizontally by 1, 2, and 4 tokens increases mean source-relative non-edit-region MSE by 0.01533, 0.01895, and 0.02304, respectively. All three settings degrade preservation (Fig.~\ref{fig:transport_evidence}(b)).
Figure~\ref{fig:transport-directional-misregistration} shows the corresponding local changes.
Shifts preserve the set of source token values while changing spatial correspondence, supporting its role in image preservation.
Circular shifts also introduce boundary wraparound; its effects and the scope of interpretation are discussed in Appendix~\ref{app:transport}.

Figure~\ref{fig:transport_evidence}(c) compares layers 20, 25, and 30 at denoising steps 0, 1, and 3 across nine experiments.
The perturbation effect of corresponding-key suppression relative to random-key suppression varies across layers and steps and weakens substantially at the final denoising step, indicating layer and temporal dependence in visual transport.
Key suppression changes the retrieved positions, whereas spatial misregistration changes the positions of condition content, testing the functional role of correspondence from two perspectives.

\paragraph{Mass-preserving spatial redistribution.}
In dual-condition Fill, we redistribute masked-image attention at step 0, layer 25 according to Eq.~\eqref{eq:redistribute}, acting only on boundary queries outside the edit region.
The zero-strength control is pixel-wise identical to the original output.
With nonzero strength, the maximum absolute deviation in total masked-image attention weight is $2.38\times10^{-7}$, while all other key weights remain unchanged.
Changes in non-edit-region and boundary MSE vary across samples, without consistent preservation gains, and overall visual changes are small (Table~\ref{tab:fill-pilot}).
This verifies that the spatial attention distribution within the masked image can be adjusted independently while preserving its total attention strength.
The effect of redistribution in the full method is evaluated by the component ablation in Table~\ref{tab:component-ablation}.

\paragraph{Mechanism analysis and method design.}
These results support decomposing text editing into the transport of reusable information and the generation of target structures.
Spatial interventions test correspondence, conditioning comparisons test content updates and appearance preservation, and spatial redistribution provides a way to independently adjust the attention distribution within a condition.
These experiments analyze local retrieval on selected samples using a single-condition model and a dual-condition reference model trained directly with text-replacement SFT.
The full method's editing quality is evaluated through component ablations that examine reconstruction initialization, the style reference, structural supervision, and spatial routing (Table~\ref{tab:component-ablation}).

\section{Conclusion}
\label{sec:conclusion}
\label{sec:discussion}
We present \method{}, a fine-grained text editing framework that combines target character generation with visual transport from the source image.
Attention Transport analyzes source detail retrieval through source weights, spatial aggregation, and feature representation quality, and strengthens information reuse in unedited regions through selective visual transport.
Multi-stage training and the Glyph-aware Structural Loss support character updates and source text appearance preservation.
We introduce TBench to systematically evaluate text updates, appearance preservation, and unedited detail retention across Alphabetic, Logographic, and Arabic numerals.
Experiments and qualitative results demonstrate the model's ability to handle length changes, dense layouts, very small text, and complex Chinese character structures.
On TBench, it achieves state-of-the-art consistency in unedited regions and full-image perceptual similarity.

\paragraph{Limitations and future work.}
Detail reuse remains limited by source image encoding quality. Original character structures in the style reference can also interfere with target character generation, causing small stroke errors or remnants of the original text.
Length changes require rearranging characters within a given edit region, and the stability of layout and appearance preservation can be improved.
Future work will explore detail encoding, disentangling appearance from character structure, and length-adaptive layouts to improve editing accuracy and stability.

\FloatBarrier
{\small
\bibliographystyle{ieeenat_fullname}
\bibliography{references}
}

\clearpage
\appendix
\section{Mechanism Derivation and Assumptions}
\label{app:derivation}
\subsection{Local Feature Aggregation Error}
Equation~\eqref{eq:localbound} assumes a reusable source feature $f(p_i)$ at the analyzed position, nonnegative normalized weights $\pi_{ij}$, and a feature field satisfying a Lipschitz condition over the aggregation range.
Since $v_j=f(p_j)+e_j$,
\begin{equation}
\begin{aligned}
 &\left\|\sum_j\pi_{ij}v_j-f(p_i)\right\|\\
 &=\left\|\sum_j\pi_{ij}[f(p_j)-f(p_i)+e_j]\right\|\\
 &\le\sum_j\pi_{ij}\|f(p_j)-f(p_i)\|
       +\sum_j\pi_{ij}\|e_j\|\\
 &\le L\sum_j\pi_{ij}\|p_j-p_i\|+\bar e_i.
\end{aligned}
\end{equation}
Sharp textures and regions spanning semantic boundaries may have a large $L$, yielding a loose bound.
The residual $e_j$ includes representation and encoding errors.
This bound separates aggregation location from condition representation in latent feature space. The full generation process additionally includes output projection, residual connections, and subsequent decoding.

Multi-axis RoPE can be written as a block rotation. Let position consist of a non-spatial identity $\tau$ and spatial coordinates $(u,v)$:
\begin{equation}
 R(p)=\operatorname{diag}
 \big(R_\tau(\tau),R_u(u),R_v(v)\big).
\end{equation}
Matching spatial positions make the latter two relative displacements zero, while different images may retain a nonzero identity displacement.
Attention depends jointly on content-dependent $q,k$ and relative position.
Spatial transformations, grids, and positional assignments together determine the correspondence that the model can learn.

\section{Implementation and Reproducibility Details}
\label{app:implementation}
\subsection{Stage 1 Full-Parameter Fine-Tuning}
\label{app:stage1}
Stage 1 uses 500,000 real-image reconstruction examples, with each original image as the target, its text region filled with black as the sole visual condition, and an instruction constructed from the original OCR text.
Initialized from Qwen-Image-Edit-2511, the model updates approximately 20.4B DiT parameters with flow matching supervision, without a style reference or Glyph-aware Structural Loss.

\begin{table}[htbp]
\centering
\caption{Training configuration for stage 1 self-supervised reconstruction.}
\label{tab:stage1-training}
\small
\begin{tabularx}{\columnwidth}{@{}lY@{}}
\toprule
Item & Setting\\
\midrule
Training examples & 500,000\\
Updated modules & All DiT parameters, $\approx$20.4B\\
Epochs / repetitions per epoch & 5 / 1\\
Training hardware & 4 nodes\\
Global batch size & 64\\
Maximum training area & $1024^2$ pixels\\
Peak learning rate & $10^{-5}$\\
Scheduler & WarmupCosineLR\\
Warmup & 500 optimization steps\\
Precision / distributed strategy & bf16 / DeepSpeed ZeRO-3\\
Gradient clipping / checkpointing & 1.0 / Enabled\\
Initialization for stage 2 & End of epoch 5\\
\bottomrule
\end{tabularx}
\end{table}

Table~\ref{tab:stage1-training} provides the full reconstruction training configuration and the checkpoint used for subsequent training.
Table~\ref{tab:stage1-base} in Sec.~\ref{sec:ablation} separately evaluates the reconstruction checkpoint at the end of epoch 1 to examine early transfer from reconstruction to text replacement.
This evaluation uses a single masked-image condition, adaptive image sizing, and a general Lightning four-step acceleration LoRA, without the task LoRA. The guidance scale is 1 and random seed is 42.
All TBench tasks used in this experiment complete generation and scoring.
OCR uses PaddleOCR 2.10.0 with PP-OCRv4 and a 10-pixel crop expansion; LPIPS uses AlexNet, and LLM scoring uses Gemini-3.0-pro.
These results correspond to inference with the four-step acceleration adapter.

\subsection{Stage 2 Data, Training, and Inference Settings}
\label{app:stage2-interface}
Stage 2 takes $(I_m,I_s,y)$ as input and uses the edited image $I_\tgt$ as the supervision target.
The target glyph support map undergoes the same spatial transformations as the target image and is used only for loss weighting.
Inference requires only a source image, edit region, and target string.

\paragraph{Training data.}
The training set contains 21,061 editing records.
All samples use masked-image, style-reference, and target-text inputs, with conditions spatially aligned to the target.

\paragraph{Optimization and resolution.}
Training inherits the fully fine-tuned stage 1 DiT; settings are listed in Table~\ref{tab:stage2-training}.
LoRA covers attention query, key, value, and output projections, as well as MLP output and modulation projections in the image and text branches.
The corresponding modules are
\texttt{to\_q}, \texttt{to\_k}, \texttt{to\_v},
\texttt{add\_q\_proj}, \texttt{add\_k\_proj}, \texttt{add\_v\_proj},
\texttt{to\_out.0}, \texttt{to\_add\_out},
\texttt{img\_mlp.net.2}, \texttt{img\_mod.1},
\texttt{txt\_mlp.net.2}, and \texttt{txt\_mod.1}.
The target image and both visual conditions use dynamic-aspect-ratio preprocessing with a maximum area of 3,500,000 pixels.
Fixed-size automatic resizing of condition images is disabled (\texttt{edit\_image\_auto\_resize=False}).
Visual conditions use a zero timestep (\texttt{zero\_cond\_t=True}).

\begin{table}[htbp]
\centering
\caption{Training configuration for stage 2 text-replacement SFT.}
\label{tab:stage2-training}
\small
\begin{tabularx}{\columnwidth}{@{}lY@{}}
\toprule
Item & Setting\\
\midrule
Initialization & Stage 1 fine-tuned DiT\\
Training records & 21,061\\
LoRA rank & 128\\
Optimizer & bitsandbytes AdamW8bit\\
Learning rate / scheduler & $2\times10^{-4}$ / ConstantLR, no warmup\\
Epochs / repetitions per epoch & 5 / 2\\
Training hardware & 4 nodes\\
Global batch size & 64\\
Maximum training area & 3,500,000 pixels, preserving aspect ratio\\
Precision / checkpointing & bf16 / Enabled\\
Structural loss weight & $\lambda_g=0.05$\\
\bottomrule
\end{tabularx}
\end{table}

\begin{table}[htbp]
\centering
\caption{Training and inference interface for stage 2.}
\label{tab:stage2-interface}
\small
\begin{tabularx}{\columnwidth}{@{}lY@{}}
\toprule
Item & Setting\\
\midrule
Initialization & Stage 1 fine-tuned DiT\\
Updates & DiT task LoRA\\
Visual conditions & masked image $I_m$, source text style reference $I_s$\\
Text condition & Target string $y$\\
Full-image supervision & Flow Matching target velocity $v^*=\epsilon-z_0$\\
Structural supervision & Local structure of predicted clean latents $\hat z_0$ and target latents $z_0$\\
Structure support map & Training-only loss weighting\\
Inference adapters & Task and four-step distillation LoRAs, both with strength 1\\
Output / VAE condition area & Both $\approx 2048^2$, preserving aspect ratio\\
Steps / guidance / seed & 4 / 1 / 42\\
Selective Visual Transport & Step 0, layer 25; four-token outer boundary; preserves total masked-image attention weight\\
\bottomrule
\end{tabularx}
\end{table}

\subsection{Glyph-aware Structural Loss Implementation}
\label{app:glyph-loss}
This section details the structural supervision in Sec.~\ref{sec:generate}. Full-image velocity supervision and local clean-latent structure supervision correspond to Eqs.~\eqref{eq:fm} and~\eqref{eq:glyph-budget}, respectively.
\paragraph{Support maps and regional alignment.}
Offline target glyph rasters undergo the same resizing and cropping as the target image, with edit boxes rounded outward to the DiT token grid.
Each DiT token corresponds to $2\times2$ unpacked latent positions, and the loss is computed within the corresponding latent region.
The glyph region is downsampled to this grid by area interpolation, normalized by its maximum to $[0,1]$, and thresholded by setting weights below 0.08 to zero.
If the pre-normalization maximum does not exceed $10^{-6}$, an all-ones regional support map is used.
Both the support map and target features are stop-gradient, while structural errors in the predicted latent backpropagate to trainable DiT parameters.

\paragraph{Gradients and amplitude normalization.}
Scharr operators are applied channel-wise to cropped latents with replicate padding. The two kernels are
\begin{equation}
 K_x=\frac{1}{32}
 \begin{bmatrix}3&0&-3\\10&0&-10\\3&0&-3\end{bmatrix},
 \qquad K_y=K_x^\top.
 \label{eq:glyph-scharr}
\end{equation}
Equation~\eqref{eq:glyph-gradient} normalizes residuals in both directions by the RMS of the target-region gradients to avoid interpreting absolute feature magnitude directly as a structural difference.
Computation uses float32, with $s_{\min}=0.02$, $\delta=0.1$, $\varepsilon_S=10^{-6}$, and $\varepsilon_R=10^{-8}$.
These constraints act on local gradient structure in latent space and do not directly define RGB stroke-orientation error.

\paragraph{Auxiliary loss budget.}
In Eq.~\eqref{eq:glyph-budget}, $\ell_\FM$ already includes the scheduler's noise weight. The adaptive coefficient is computed per sample and detached from gradients.
We set $\lambda_g=0.05$, limiting the auxiliary loss value to at most 5\% of the sample's base loss; noise gating can further reduce this fraction.
Structural support weights affect only the auxiliary term. Base Flow Matching still averages uniformly over the full latent.

\subsection{Spatial Redistribution and Mechanism Analysis Settings}
\label{app:mechanism-settings}
\method{} applies spatial redistribution at step 0, layer 25 to queries in a four-token-wide band outside the edit box.
For an edit box whose longer side spans at most 4, more than 4 and at most 10, or more than 10 tokens, we use $(r,\beta)=(2,0.15),(4,0.25),(6,0.35)$, respectively.
The table below specifies the dual-condition Fill reference configuration used for mechanism analysis in Sec.~\ref{sec:causal}.
It performs text-replacement SFT directly on the original backbone, without stage 1 reconstruction initialization or the Glyph-aware Structural Loss.
Its spatial redistribution rule matches the full method.

\begin{table}[htbp]
\centering
\caption{Training and inference settings of the Fill reference configuration for mechanism analysis.}
\small
\begin{tabular}{ll}
\toprule
Item & Setting\\
\midrule
Backbone & Qwen-Image-Edit-2511\\
Fine-tuning & DiT LoRA, rank 128\\
Training records & 21,061\\
Maximum training area & 3,500,000 pixels\\
Learning rate / scheduler & $2\times10^{-4}$ / ConstantLR\\
Epochs / repetitions per epoch & 5 / 2\\
Training hardware & 4 nodes\\
Global batch size & 64\\
Checkpoint & End of epoch 4\\
Output / condition target area & $2048^2$ / $2048^2$\\
Sampling steps / guidance & 4 / 1\\
LoRA strengths / seed & Both 1 / 42\\
\bottomrule
\end{tabular}
\end{table}

\paragraph{Condition construction.}
The masked image fills the edit box with black. The style reference retains the source content inside that box at its original position on an equally sized black canvas.
Condition images preserve dynamic aspect ratios, with training and inference areas set by each configuration.
To blank the style reference, we use an equally sized black image, preserving the image count and token layout while changing both VAE and VLM encodings.
A black reference may deviate from the training input distribution.
Figure~\ref{fig:fill-condition-roles} presents an additional paired experiment on eight source images under the same configuration.
It compares outputs with and without the style reference at fixed target text, and outputs for two target strings at fixed visual conditions; other inputs, model weights, and random seeds remain fixed.
These eight source images form the qualitative paired set. Attention and routing statistics use the samples described below.

\paragraph{Attention collection.}
Attention measurements and spatial redistribution experiments use TBench samples, with model weights, source image, edit box, and random seed fixed for each example.
For the original Fill model, we record combinations of denoising steps 0 and 3 with Transformer layers 6, 25, and 54.
Target replacement, reference blanking, and nonzero redistribution are recorded at layer 25 for steps 0 and 3. Layers and steps are zero-indexed.
Each combination samples a fixed $24\times24$ query grid across the full image and all queries in the edit region and the four-token-wide outer boundary.
Figure~\ref{fig:fill-mechanism} uses a contiguous local grid at step 0, layer 25.
Figure~\ref{fig:spatial-kernel} reuses the original queries and keys saved at this location for these samples.

We compute source weights and per-head weighted aggregated features for both conditions using Eq.~\eqref{eq:transport}.
Figure~\ref{fig:fill-mechanism} shows channel RMS after concatenating head features, with a shared color scale for both conditions.
Measurements are taken before the attention output projection and gating; RMS describes the magnitude of weighted aggregated features.
Source weights are averaged first over attention heads, then over regional queries.
In Table~\ref{tab:fill-pilot}, $\bar\mu_e^{(s)}$ and $\bar\mu_b^{(s)}$ denote the mean total attention weights assigned to the style reference by the edit region and outer boundary, respectively.

\paragraph{Redistribution implementation and statistics.}
Interventions use the layer, step, and $(r,\beta)$ settings specified in this section, without retuning them for the measurement samples.
The implementation adds the weighted feature difference $\Delta h_i=\sum_{j\in\mathcal K_m}(a'_{ij}-a_{ij})v_j$ to the native fused-attention output.
At zero strength, $\Delta h_i$ is explicitly set to zero, and saved images are pixel-wise identical to the original outputs.
For nonzero interventions, the maximum absolute deviation in total masked-image attention weight is $2.38\times10^{-7}$, while all other key weights remain unchanged.

The source image is mapped to the output dimensions using bicubic interpolation, with RGB normalized to $[0,1]$.
Edit-box endpoints are scaled and rounded down, using a half-open interval that excludes the bottom-right endpoint. The boundary is a 64-pixel-wide rectangular ring outside the box.
Let $E_u^{(0)}$ and $E_u^{(r)}$ denote source-relative regional MSE for the original and redistributed outputs.
Table~\ref{tab:fill-pilot} reports $\delta_u=100(E_u^{(r)}/E_u^{(0)}-1)$, where $u=n,b$ denotes the non-edit region and boundary, respectively.

\begin{table}[htbp]
\centering
\caption{\textbf{Condition attention weights and spatial redistribution.}
Each row represents one measured sample, with all values expressed as percentages.
$\bar\mu^{(s)}$ is the mean total attention weight assigned to the style reference by the original Fill model.
$\delta_n$ and $\delta_b$ are changes in non-edit-region and boundary MSE after redistribution relative to the original output; negative values indicate closer agreement with the source image.
}
\label{tab:fill-pilot}
\small
\setlength{\tabcolsep}{4pt}
\begin{tabular}{lrrrr}
\toprule
Sample & $\bar\mu^{(s)}_e$ & $\bar\mu^{(s)}_b$ & $\delta_n$ & $\delta_b$\\
\midrule
Sample 1 & 10.63 & 6.26 & -0.004 & -0.760\\
Sample 2 & 10.37 & 5.91 & +0.035 & +0.210\\
Sample 3 & 11.92 & 6.79 & -0.159 & +1.012\\
Sample 4 & 13.31 & 6.07 & +1.508 & +1.405\\
\bottomrule
\end{tabular}

\end{table}

\paragraph{Target text and glyph inputs.}
\label{app:glyph_input}
Text replacement uses the masked image $I_m$ and style reference $I_s$ constructed from the source image; stage 1 reconstruction uses only $I_m$.
The target string $y$ is supplied through a natural-language instruction.
During training, the target image provides flow matching supervision, and the glyph support map defines only spatial weights for the structural loss.
Inference requires no rendered target glyphs.
The style reference provides source text appearance, while target text and replacement supervision jointly constrain new characters.

\section{Condition Construction and Evaluation Implementation}
\label{app:region}
\paragraph{Image and coordinate processing.}
Annotation coordinates are first mapped according to the size ratio between the annotated image and source image, then transformed to the output through image resizing and cropping.
The masked image uses Pillow rectangle filling, which includes the bottom-right endpoint.
Style crops and evaluation masks use half-open intervals excluding that endpoint, so boundaries may differ by one pixel.
If annotations lack valid image dimensions, coordinate mapping uses the actual source image dimensions.
Reference images are mapped to the output using known size transformations.
SSIM first computes a local similarity map between the source and output, then averages over window centers in the non-edit region; windows may cross the edit boundary.
LPIPS is computed over the full image.

\paragraph{Reference images and LLM evaluation.}
Each evaluation task uses its fixed reference image and evaluation template.
Gemini-3.0-pro provides LLM scores from the source and generated images, both resized to a longest side of 1024 before encoding.
LLM and human evaluation results are reported in Tables~\ref{tab:tbench} and~\ref{tab:human-review}, respectively.

\section{Paired Comparison on TBench}
\label{app:comparison}
Both configurations are evaluated using a single image marked by a red box.
Following the order in the reference model comparison in Sec.~\ref{sec:ablation}, the first trains a Fill LoRA on the original backbone, while the second trains a single-image editing LoRA on the fully fine-tuned backbone.
They use different training data and resolution settings.

TBench results are paired by source image, with 10,000 bootstrap resamples and random seed 42.
Differences in Table~\ref{tab:paired} are the first configuration minus the second in Sec.~\ref{sec:ablation}, characterizing the trade-off between text accuracy and image preservation.
Intervals capture image-resampling variability, excluding variability from generation seeds, scoring services, or training randomness.

\begin{table}[htbp]
\centering
\caption{Paired differences and 95\% bootstrap intervals on TBench. OCR differences are expressed as proportions.}
\label{tab:paired}
\small
\setlength{\tabcolsep}{3pt}
\begin{tabular}{lrr}
\toprule
Metric & Difference & 95\% interval\\
\midrule
OCR$\uparrow$ & $-.1055$ & $[-.1658,-.0452]$\\
ED$\downarrow$ & $+.0933$ & $[+.0526,+.1363]$\\
Non-edit SSIM$\uparrow$ & $+.0733$ & $[+.0651,+.0820]$\\
Full-image LPIPS$\downarrow$ & $-.0818$ & $[-.0894,-.0742]$\\
LLM$\uparrow$ & $-.1960$ & $[-.3920,.0000]$\\
\bottomrule
\end{tabular}
\end{table}

\section{Attention Transport Intervention Settings}
\label{app:transport}
A single-condition editing model tests the functional role of spatial correspondence.
It shares the backbone of the two-stage method but uses independently trained weights and a single visual condition.
All interventions fix model weights, conditions, and random states.

\paragraph{Local key suppression.}
On TBench images, we suppress condition-key logits for non-edit queries at layer 30, denoising step 0.
The corresponding-key group uses a neighborhood radius of 8 tokens, suppressing $17\times17=289$ keys per query with strength 50.
The random-key group matches the number of keys, layer, step, and strength.

In Fig.~\ref{fig:transport_evidence}(a), 90\% of samples lie above the identity line. The mean of per-sample MSE ratios is 1.856, distinct from the ratio of the two group means.
RGB values are normalized to $[0,1]$.
The random-key group does not match the total suppressed attention weight, so sensitivity to high-weight keys may also affect the results.

\begin{figure*}[tp]
\centering
\includegraphics[width=\textwidth]{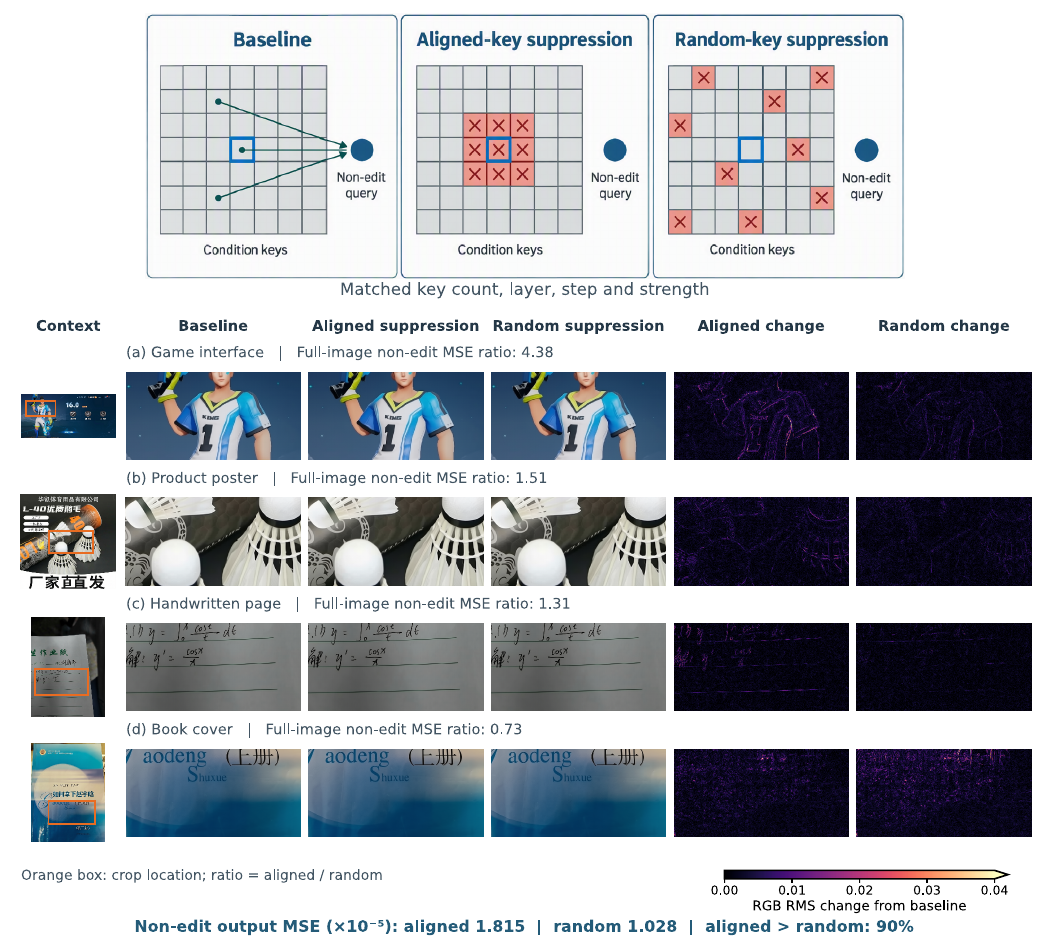}
\caption{\textbf{Additional visualizations of local key suppression.}
Interventions act at layer 30, denoising step 0.
Top: suppression of locally corresponding or random keys for non-edit queries, matching key count, layer, step, and suppression strength.
Bottom: a game interface, product poster, handwritten page, and book cover illustrate strong, near-median, and reversed intervention effects.
Orange boxes in the left thumbnails mark enlarged non-edit regions. Subsequent columns show the baseline, corresponding-key suppression, random-key suppression, and RGB root-mean-square differences of both interventions from the baseline.
Each row reports the corresponding-key/random-key perturbation MSE ratio over all non-edit regions in the full image. Differences are computed at native resolution with a shared color scale.
The bottom statistics show that corresponding-key suppression causes larger perturbations in 90\% of samples.
Layer and step statistics are shown in Fig.~\ref{fig:transport_evidence}(c).
}
\label{fig:transport-directional-suppression}
\end{figure*}

Figure~\ref{fig:transport-directional-suppression} provides additional local outputs and differences from the unmodified outputs.

\paragraph{Condition latent misregistration.}
On TBench images, the misregistration experiment circularly shifts the condition latent grid horizontally while keeping positional assignments and other inference settings fixed.
Figure~\ref{fig:transport_evidence}(b) reports increased source-relative error in unedited regions; all three settings degrade preservation.

Mean degradation increases with displacement.
Circular shifts preserve the set of token values while changing spatial positions and introducing boundary wraparound.
Circular shifts and local key suppression differ in intervention strength; each experiment compares its own intervention with its control.

\paragraph{Paired comparisons across layers and denoising steps.}
Figure~\ref{fig:transport_evidence}(c) summarizes the ratios of the two groups' mean MSE across nine experiments.
Cell values show ratios of mean MSE, with a logarithmic color scale.
Only evaluated layers and steps are shown.

\begin{table*}[t]
\centering
\caption{Paired intervention effects with a single visual condition.
The first row compares perturbations relative to unmodified outputs; the remaining rows compare increases in source-relative non-edit-region MSE. The two error references are not directly comparable.
Intervals are obtained by paired image-wise bootstrap.
}
\label{tab:mechanism-paired}
\small
\begin{tabular}{lrrr}
\toprule
Intervention contrast & Mean difference & 95\% CI & Positive differences\\
\midrule
Corresponding-key suppression $-$ random-key suppression & $7.867\times10^{-6}$ & $[4.662,11.510]\times10^{-6}$ & 90\%\\
Condition shift of 1 token $-$ no shift & 0.01533 & $[0.01069,0.02034]$ & All positive\\
Condition shift of 2 tokens $-$ no shift & 0.01895 & $[0.01357,0.02459]$ & All positive\\
Condition shift of 4 tokens $-$ no shift & 0.02304 & $[0.01683,0.02932]$ & All positive\\
\bottomrule
\end{tabular}
\end{table*}

\paragraph{Effect uncertainty.}
Table~\ref{tab:mechanism-paired} uses 10,000 paired image-level bootstrap resamples with random seed 42, taking the 2.5\% and 97.5\% quantiles of the mean distribution as interval endpoints.
Each resample preserves intervention--control pairs.
Intervals capture variability within the current image set, excluding retraining or generation-seed variability.

\begin{figure*}[tp]
\centering
\includegraphics[width=.98\textwidth]{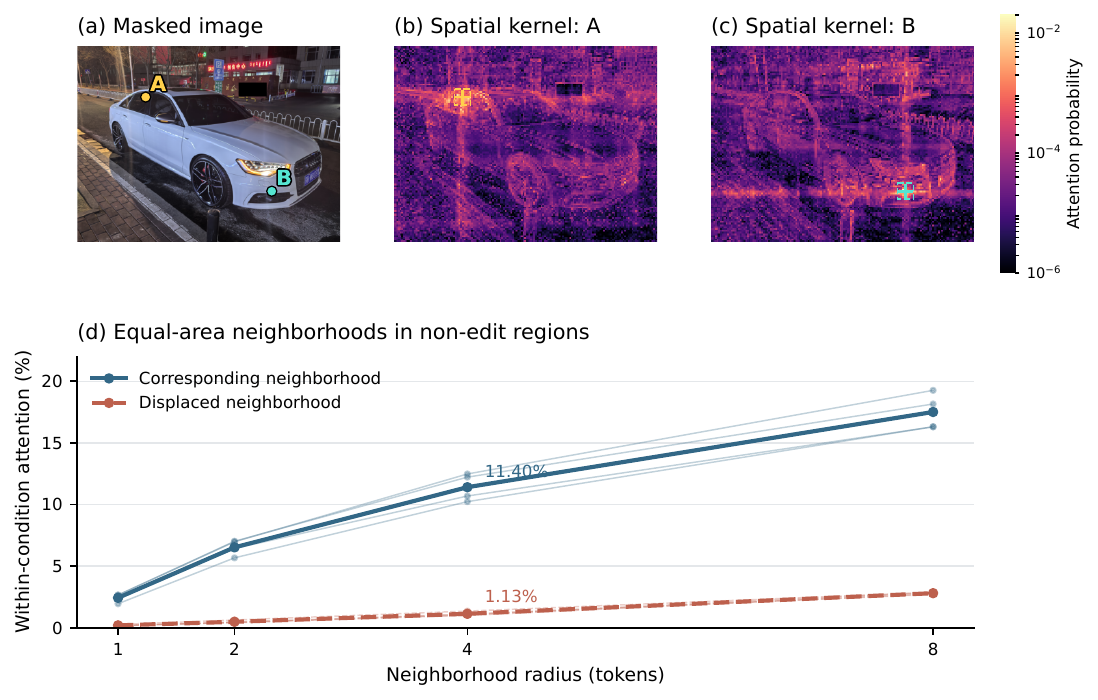}
\caption{\textbf{Spatial attention distributions within a condition.}
(a) The masked image with two non-edit query positions, A and B, marked by spatial correspondence.
(b,c) Spatial aggregation kernels over the masked image for the two queries. Each pixel corresponds to a condition key position; crosses mark corresponding positions, and dashed boxes show radius-4-token neighborhoods. Both maps use the same logarithmic color scale.
(d) Attention fractions in corresponding and equally sized displaced neighborhoods. Thin lines represent individual samples, and thick lines show sample means.
Measurements use dual-condition Fill at denoising step 0, Transformer layer 25.
Attention is normalized within the masked image per head, then averaged across heads. Layers and steps are zero-indexed.
}
\label{fig:spatial-kernel}
\end{figure*}

\section{Spatial Attention Distributions Within a Condition}
\label{app:feature-maps}
To directly inspect where the model retrieves information from a condition, we reuse the original Q/K from the dual-condition Fill configuration in Appendix~\ref{app:mechanism-settings}.
We analyze the spatial attention distribution of output queries over the masked image at denoising step 0, Transformer layer 25.
For output query $i$ and attention head $h$, normalization within the masked-image key set $\mathcal K_m$ gives
\begin{equation}
\begin{aligned}
 \pi^{(h,m)}_{ij}&=\frac{\exp(q_i^{(h)\top}k_j^{(h)}/\sqrt d)}
 {\sum_{k\in\mathcal K_m}\exp(q_i^{(h)\top}k_k^{(h)}/\sqrt d)},\\
 \bar\pi^{(m)}_{ij}&=\frac{1}{N_h}\sum_h\pi^{(h,m)}_{ij}.
\end{aligned}
\end{equation}
For a fixed query $i$, we arrange $\bar\pi^{(m)}_{ij}$ by the spatial positions of condition keys $j$.
The distribution sums to 1 and describes retrieval locations within the masked image, excluding its total attention weight.

\paragraph{Query positions and spatial retrieval.}
Figure~\ref{fig:spatial-kernel}(a--c) uses the same sign scene as Fig.~\ref{fig:fill-mechanism}.
We choose two valid non-edit queries on the fixed grid closest to the positions at $1/4$ and $3/4$ of both the image width and height.
As the query moves from A to B, stronger attention responses move toward the corresponding condition neighborhood, illustrating spatial correspondence between output positions and source retrieval locations.

\paragraph{Corresponding and displaced neighborhoods.}
We further compute statistics over non-edit regions on the fixed $24\times24$ query grid.
Since the output and masked image share a latent grid, the corresponding key position for query $i$ has the same grid coordinates.
We define a square neighborhood $\mathcal N_r(i)$ of Chebyshev radius $r$ centered at this position, with attention fraction
\begin{equation}
 C_i(r)=\sum_{j\in\mathcal N_r(i)}\bar\pi^{(m)}_{ij},
 \qquad r\in\{1,2,4,8\}.
\end{equation}
The displaced control shifts neighborhood centers horizontally or vertically by approximately one-third of the grid width or height, retains valid neighborhoods entirely within non-edit regions, and averages their attention fractions.
Both groups contain $(2r+1)^2$ keys per neighborhood, without boundary clipping or circular wraparound.
All radii use the same query set. At the maximum radius of 8, neighborhoods touching the edit region or image edges are excluded, leaving 1,463 queries.
We first average over queries within each image, then average equally across images.

In Fig.~\ref{fig:spatial-kernel}(d), corresponding neighborhoods receive more attention than displaced neighborhoods at every tested radius.
For $r=4$, their mean attention fractions are 11.40\% and 1.13\%, respectively, a ratio of approximately 10.06.
Under this configuration, spatial aggregation within the masked image favors neighborhoods corresponding to output positions, providing distributional evidence of local source retrieval.
Its functional effects on generation are assessed together with the interventions in Figs.~\ref{fig:transport_evidence} and~\ref{fig:transport-directional-misregistration}.

\section{Statistical Definitions for Output Change Distributions}
\label{app:change-analysis}
For output $\hat I$ and source image $\tilde I_\src$ mapped to the output dimensions, we define per-pixel squared change as
\begin{equation}
 D(p)=\frac{1}{3}\sum_{c=1}^{3}
   [\hat I(p,c)-\tilde I_\src(p,c)]^2.
\end{equation}
Figure~\ref{fig:region-change}(a) reports $C_u=|\Omega_u|^{-1}\sum_{p\in\Omega_u}D(p)$ for $u\in\{e,n\}$.
Panel (b) compares edit-region area fraction $\rho_e$ with its share of total squared change:
\begin{equation}
 \eta_e=\frac{\sum_{p\in\Omega_e}D(p)}{\sum_{p\in\Omega}D(p)}.
\end{equation}
$C_e$ measures content changes in the edit region, and $\eta_e$ describes the spatial concentration of changes. Target character correctness is assessed by text accuracy metrics.

This analysis uses TBench outputs from the dual-condition Fill reference configuration in Appendix~\ref{app:mechanism-settings}, with actual generated dimensions and RGB normalized to $[0,1]$.
Source images are mapped to the output grid by bicubic interpolation.
Output dimensions follow aspect-ratio-preserving area scaling and rounding to multiples of 32.

Edit boxes use original-image coordinates from the fixed evaluation annotation list and are scaled to the output dimensions.
Top-left endpoints are rounded down and bottom-right endpoints up, with half-open intervals excluding the latter.
Change statistics follow the definitions in this section. Figure~\ref{fig:region-change}(c) uses per-image OCR decisions and SSIM scores.

\clearpage
\balance
\section{Parameter Adaptation Analysis}
\label{app:training-analysis}
\begin{figure}[htbp]
\centering
\includegraphics[width=\linewidth]{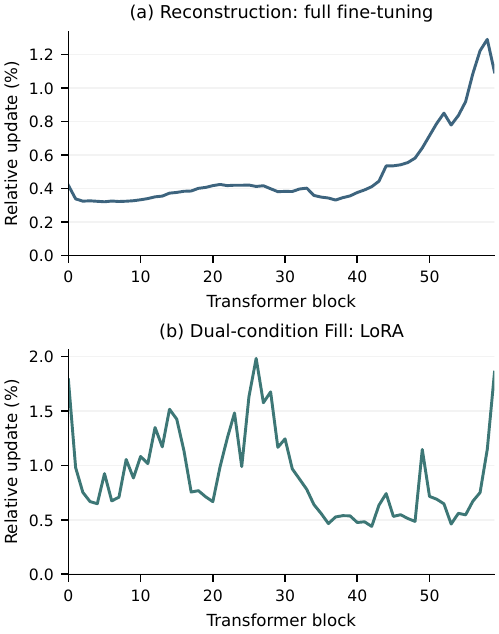}
\caption{\textbf{Layer-wise relative parameter updates.}
Reconstruction includes all parameters in each layer, whereas Fill includes projection modules covered by LoRA. Updates are normalized by the Frobenius norm of the corresponding initial parameters.
Layer indices start at 0.}
\label{fig:parameter-update-distribution}
\end{figure}

We analyze where parameters adapt during self-supervised text-region reconstruction and training of the dual-condition Fill reference model.
For reconstruction, we use the full-parameter checkpoint at the end of epoch 1; for Fill, we use rank-128 LoRA weights at the end of epoch 4, with the reference configuration in Appendix~\ref{app:mechanism-settings}.
The two analyses characterize update distributions across layers under their respective configurations, which differ in training settings, checkpoint epochs, and parameter coverage.

For a parameter group $G$ within a layer, we define the relative update as
\begin{equation}
 R_G=\left(
 \frac{\sum_{m\in G}\|\Delta W_m\|_F^2}
      {\sum_{m\in G}\|W_m^{(0)}\|_F^2}
 \right)^{1/2}.
 \label{eq:relative-parameter-update}
\end{equation}
Here, $W_m^{(0)}$ denotes the initial parameters.
For full-parameter fine-tuning, $\Delta W_m=W_m-W_m^{(0)}$.
For LoRA, $\Delta W_m=(\alpha/r)B_mA_m$, with $\alpha=r=128$ in this configuration.

\begin{samepage}
Figure~\ref{fig:parameter-update-distribution} shows larger relative updates in later layers for reconstruction, reaching 1.29\% at layer 58.
For Fill, larger updates occur in the middle layers and at both ends of the network.
Relative updates characterize the locations and magnitudes of parameter adaptation, while interventions and ablations evaluate the functional roles of the modules.
\par
\end{samepage}

\end{document}